\documentclass[sigconf]{acmart}
\usepackage{cleveref}
\usepackage{enumitem} 
\usepackage{multirow}
\usepackage{array}
\usepackage{hyperref}
\usepackage{academicons}
\usepackage{xcolor}

\usepackage{fancyhdr}
\newcommand{\changefont}{%
    \fontsize{7}{11}\selectfont
}
\AtBeginDocument{%
  \providecommand\BibTeX{{%
    \normalfont B\kern-0.5em{\scshape i\kern-0.25em b}\kern-0.8em\TeX}}}

\copyrightyear{2022}
\acmYear{2022}
\setcopyright{acmlicensed}
\acmConference[MM '22] {Proceedings of the 30th ACM International Conference on Multimedia}{October 10--14, 2022}{Lisboa, Portugal.}
\acmBooktitle{Proceedings of the 30th ACM International Conference on Multimedia (MM '22), Oct. 10--14, 2022, Lisboa, Portugal}
\acmPrice{15.00}
\acmISBN{978-1-4503-9203-7/22/10}
\acmDOI{10.1145/3503161.3547821}

\newcommand{\etal}{\textit{et al}.}
\newcommand{\etc}{\textit{etc}}
\newcommand{\ie}{\textit{i.e.}}
\newcommand{\eg}{\textit{e.g.}}

\newcommand{\aka}{\textit{a.k.a}}

\begin{document}

\title{Adma-GAN: Attribute-Driven Memory Augmented GANs for Text-to-Image Generation.}


\author{Xintian Wu}
\orcid{0000-0002-2988-5215}
\affiliation{%
  \institution{Zhejiang University}
  \city{Hangzhou}
  \country{China}
}
\email{hsintien@zju.edu.cn}
\author{Hanbin Zhao}
\orcid{0000-0001-8906-4534}
\affiliation{%
  \institution{Zhejiang University}
  \city{Hangzhou}
  \country{China}
}
\email{zhaohanbin@zju.edu.cn}
\author{Liangli Zheng}
\affiliation{%
  \institution{Zhejiang University}
  \city{Hangzhou}
  \country{China}
}
\email{lianglizheng@zju.edu.cn}
\orcid{0000-0002-3750-7150}
\author{Shouhong Ding}
\orcid{0000-0002-3175-3553}
\affiliation{%
  \institution{Youtu Lab, Tencent}
  \city{Shanghai}
  \country{China}
}
\email{ericshding@tencent.com}
\author{Xi Li}
\orcid{0000-0003-3023-1662}
\authornote{Corresponding Author}
\affiliation{%
  \institution{Zhejiang University}
  \city{Hangzhou}
  \country{China}
}
\additionalaffiliation{
  \institution{Shanghai AI Laboratory}
  \city{Shanghai}
  \country{China}
}
\email{xilizju@zju.edu.cn}

\renewcommand{\shortauthors}{Xintian Wu et al.}

\begin{abstract}
  
  As a challenging task, text-to-image generation aims to generate photo-realistic and semantically consistent images according to the given text descriptions. Existing methods mainly extract the text information from only one sentence to represent an image and the text representation effects the quality of the generated image well. 
  However, directly utilizing the limited information in one sentence misses some key attribute descriptions, which are the crucial factors to describe an image accurately.
  To alleviate the above problem, we propose an effective text representation method with the complements of attribute information. Firstly, we construct an attribute memory to jointly control the text-to-image generation with sentence input. Secondly, we explore two update mechanisms, sample-aware and sample-joint mechanisms, to dynamically optimize a generalized attribute memory. Furthermore, we design an attribute-sentence-joint conditional generator learning scheme to align the feature embeddings among multiple representations, which promotes the cross-modal network training. Experimental results illustrate that the proposed method obtains substantial performance improvements on both the CUB (FID from 14.81 to 8.57) and COCO (FID from 21.42 to 12.39) datasets.

\end{abstract}



  



\maketitle

\section{Introduction}

\begin{figure}[!t]
	\centering
	\includegraphics[width=\linewidth]{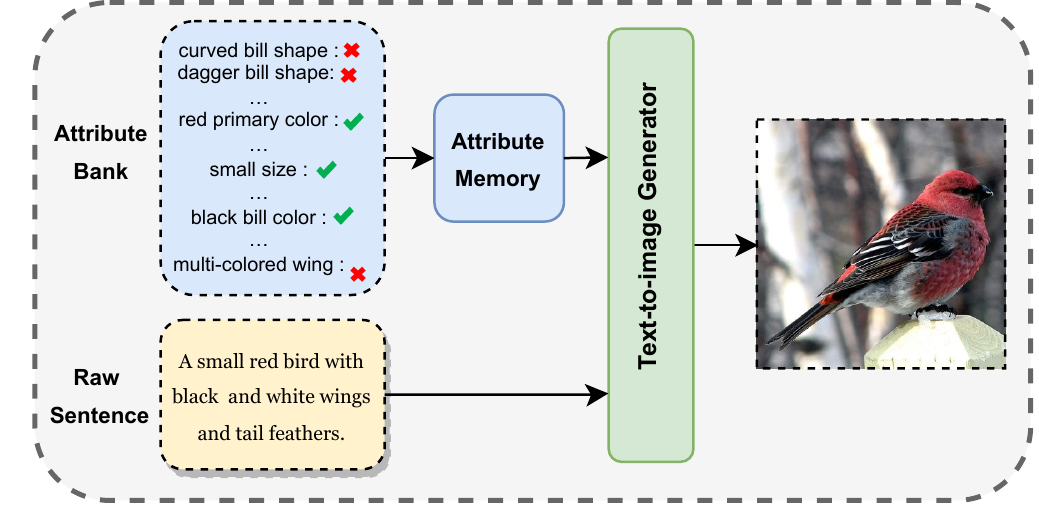}
	\caption{An example of explanation and reasoning in the proposed text-to-image generation framework. A bird image is annotated with a sentence and a multi-attribute label in CUB dataset.}
	\label{fig:motivation}
  \vspace{-1.5em}
\end{figure}

Multimodal data~\cite{wu2021mgh,srivastava2012learning} has been widely used in numerous cross-modal tasks~\cite{antol2015vqa,xiao2021boundary,yang2021deconfounded,chen2017show,zhang2017stackgan}. Among them, text-to-image generation is a research hotspot, which
aims to generate photo-realistic images according to the text description and has great potential for applications \eg, image editing~\cite{wu2021n,wu2021f3a}, computer-aided design, entertainment interaction~\cite{nvidia2021gaugan2}. 
Due to the large gap of modal structure between text and image data, the optimization of the cross-modal text-to-image generative models is prone to overfitting or collapse, resulting in generating irregular object shape. 


In the literature, most of the methods formulate the task with a conditional GAN formulation~\cite{mirza2014conditional}. They usually first embedded the text description, and then set it as the conditional input of the generator for image generation.
Typically, the text representation effects the quality of the generated image well. Existing sentence embedding methods only utilize the limited information in one sentence to model the text representation, resulting in the following obstacle: 
it misses some key attribute descriptions, which are the crucial factors to describe an image accurately.
To alleviate it, we propose an effective text representation method with the complements of attribute information of the sentence. 
As shown in~\cref{fig:motivation}, we design an attribute memory to jointly control the generator with sentence. 
Hence, the text-to-image generation task can be considered as an attribute-sentence-joint conditional generation problem. The key points of this problem lie in two aspects: 1) how to construct an attribute memory, 2) how to learn the conditional image generator with the joint attribute and sentence conditions.

As for the attribute memory construction, we first collect all the possible attribute descriptions within a dataset as an attribute bank and convert them into an attribute memory. Then, we use the attribute label to extract the corresponding embeddings from the attribute memory and combine them into a common attribute embedding as the condition. Considering that attribute descriptions and images are two different modalities, a fixed memory is difficult to achieve cross-modal image generation. Therefore, we design a learnable memory update scheme to obtain an optimal attribute memory. Specifically, we explore two different mechanisms for optimization, the sample-aware and sample-joint update mechanisms. For the sample-aware one, we treat each sample in isolation and average all the sample-aware attribute embeddings together. In this way, only sample-related attribute embeddings will be optimized in one training session through gradient back-propagation.
However, co-occurrence exists between attributes and attribute pairs are combined differently in different samples. 
The sample-aware mechanism only considers the co-occurrence inside an image while ignores the global correlation patterns within the entire dataset. Therefore, we further propose to model the sample-joint attribute relationship. Specifically, we construct a graph to represent the attribute correlation within a dataset and use GCNs~\cite{kipf2016semi,chen2019multi} to extract attribute features. Thus, the co-occurrence attribute embeddings will be updated with edge connections to get a more suitable one.
Experimental results show that the improved sample-joint strategy maps multiple attributes to the images better.


After building the attribute memory, an attribute-sentence-joint conditional generator learning scheme is designed to handle the conversion among multiple representations (\ie, sentence, attribute, image).
In our scheme, an image should strictly correspond to both the sentence and attribute. Therefore, we propose to align the image with sentence and attribute in a common space using contrastive learning. Attribute-image and sentence-image pairs belonging to the same sample are pulled closer, while pairs of different samples are pushed farther. 

Combining the above strategies, we improve DF-GAN~\cite{tao2020df} and obtain substantial performance improvements on the CUB~\cite{welinder2010caltech} and COCO~\cite{lin2014microsoft} datasets. 
Our contributions can be concluded as follow:

\begin{itemize}[leftmargin=*]
  \item We propose a novel attribute representation as an additional condition and construct an attribute memory to augment the text-to-image generation with richer information. 
  \item We design a memory update scheme, including a sample-aware and a sample-joint update mechanism to obtain the optimal attribute memory for attribute-driven conditional generation. 
  \item We introduce contrastive learning to enhance semantic consistency among multiple representations, which facilitates the training of cross-modal GAN well. 
\end{itemize}
\section{Related Work}





\begin{figure*}[!t]
	\centering
	\includegraphics[width=0.95\linewidth]{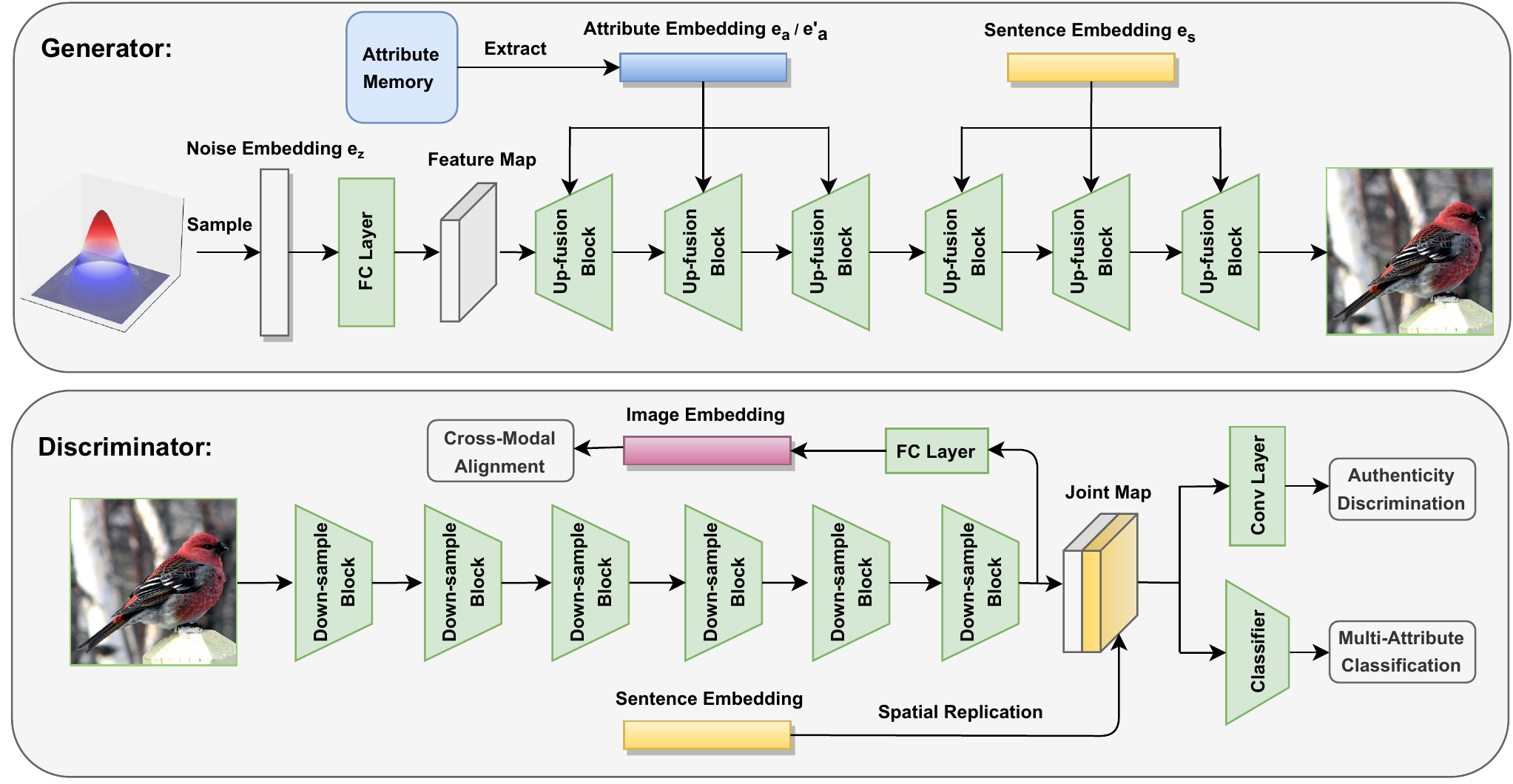}
	\caption{Illustration of the proposed text-to-image generation framework.}
	\label{fig:framework}
\end{figure*}

\subsection{Text Representation}

Generating an image from a single sentence is a process creating information from less to more, which is difficult to optimize the text-to-image generation model in practice. To alleviate such a problem, many works are devoted to enrich the text representation. 1) \textit{Providing additional descriptions}. Cheng \etal~\cite{cheng2020rifegan} and Sun \etal~\cite{sun2021multi} combined multiple captions together instead of one sentence input, and EI \etal~\cite{el2019gilt} proposed a novel task of synthesizing images from long text. Sharma \etal~\cite{sharma2018chatpainter} and Frolov \etal~\cite{frolov2020leveraging} introduced the dialog and questions.
These additional descriptions provide richer content to reduce the gap across modalities. 2) \textit{Mining more representations from one sentence}. Xu \etal~\cite{xu2018attngan} used word features to make the network pay attention to word-level image-text matching. Han \etal~\cite{han2020victr} disassembled the subject-predicate-object structure of the sentence to generate scene graph embedding. Ruan \etal~\cite{ruan2021dae} proposed to extract the aspect information from one sentence. Zhu \etal~\cite{zhu2020cookgan} modeled the instructions and ingredients of the dish description separately in the Recipe1M dataset~\cite{salvador2017learning}. In this paper, we propose a novel attribute representation, which complements the text information of a single sentence, and enables the network to focus more on the key descriptions. 



\subsection{Text-to-image Network Design}

A powerful network architecture is also a key to the generation task. Several works focus on designing better network architecture for text-to-image generation. 1) \textit{Progressive generation}. 
Zhang \etal~\cite{zhang2017stackgan} achieved cross-modal generation by first generating an initial image and then refining it to a larger resolution. There are many follow-up works~\cite{zhang2018stackgan++,zhang2018photographic,kim2020tivgan} further optimized the structure based on such an idea. In addition, Li \etal~\cite{li2019object} and Hong \etal~\cite{hong2018inferring} proposed to convert text into bounding box first, then into label map, and finally performed image-to-image translation to achieve transformation. 
2) \textit{Cross-modal fusion}. Most of the works introduce attention mechanism to fuse the cross-modal representations. Xu \etal~\cite{xu2018attngan} proposed AttnGAN, allowing attention-driven refinement for fine-grained text-to-image generation. Many follow-up works attempted to make improvement, including dual attention~\cite{cai2019dualattn}, segmentation attention~\cite{gou2020segattngan}, memory attention~\cite{zhu2019dm,li2021memory} \etc. Tao \etal~\cite{tao2020df} further proposed a simple but more efficient fusion method, named DF-GAN, to synthesize realistic and text-matching images. 
In this paper, we improve the network structure based on DF-GAN with a learnable attribute memory, and design an attribute-sentence-joint conditional framework for text-to-image generation.

\subsection{Cross-modal Training Optimization}

There exists another aspect to improve the model performance, \ie, optimizing the network training. The core optimization directions can be divided into the following three parts. 1) \textit{Large-scale pretrained model}. 
Ramesh \etal~\cite{ramesh2021zero} and Ding \etal~\cite{ding2021cogview} have proposed billion-parameter-level models and million-level datasets for training, both of which achieved amazing effect in society. Wu \etal~\cite{wu2021n} even extended this task to text-to-video generation with large-scale training strategies. 
2) \textit{Cycle consistency}. Qiao \etal~\cite{qiao2019mirrorgan} proposed a text-to-image-to-text framework that utilizes bidirectional generation to maintain semantic consistency. Wang \etal~\cite{wang2021cycle} utilized GAN-inversion~\cite{xia2021gan} technology to search the corresponding text embedding in reverse to optimize the text encoder. 3) \textit{Contrastive learning}. Yin \etal~\cite{yin2019semantics} and Ye \etal~\cite{ye2021improving} applied contrastive learning to constrain similar sentence for generating consistent images. Zhang \etal~\cite{zhang2021cross} constructed cross-modal pairs, achieving remarkable results. Furthermore, we apply contrastive learning to align multiple representations including attribute, sentence and image, and thus improves the model performance well.  
\section{Method}


\subsection{Overview}\label{sec:overview}
Traditional text-to-image generation approaches learn a mapping function $\mathcal{F}_{trad}$ to transform a noise space $\mathcal{Z}$ into an image space $\mathcal{X}$ conditioned on the sentence space $\mathcal{S}$, \ie, $\mathcal{F}_{trad}: (\mathcal{Z}|\mathcal{S})\rightarrow \mathcal{X}$. However, utilizing the limited information in one sentence is not enough to accurately describe an image. Therefore, we introduce an attribute label space $\mathcal{Y}$, and propose a framework to generate images from both attribute and sentence representations. 
It is thus an attribute-sentence-joint conditional GAN pipeline, learning a mapping function $\mathcal{F}_{adma}: (\mathcal{Z}|\mathcal{S},\mathcal{Y})\rightarrow \mathcal{X}$. 
The proposed framework includes a memory augmented generator, and a conditional discriminator with auxiliary classification. 

As shown in~\cref{fig:framework}, the generator is conditioned on the sentence embedding $e_{s}$ and attribute embedding $e_{a}$. We first convert the raw sentence into sentence embedding $e_{s}$ and construct an attribute memory $M_{a}$ from a pre-defined attribute bank. Then, attribute embedding $e_{a}$ is extracted from $M_{a}$ using a multi-attribute label $y$. After obtaining $e_{a}$ and $e_{s}$, we insert them into the low-level blocks and high-level blocks respectively to modulate features at different levels. Each block is a ResBlock~\cite{he2016deep} for text-image fusion and resolution enhancement. 


The discriminator aims to determine the real probability and attribute probability of a real image $x$ or generated image $x_{f}$. It is composed of several down-sample ResBlocks followed by two fully connected branches. We first extract the input image into low-dimensional map, and then concatenate the sentence embedding on it as the joint map. Finally, the joint map is sent into two different layers, one for authenticity discrimination and the other one for multi-attribute classification. In this way, the discriminator is trained to obtain the discriminability and classification ability, and thus provides better optimization directions for the generator.

\begin{figure*}[!t]
	\centering
	\includegraphics[width=0.98\linewidth]{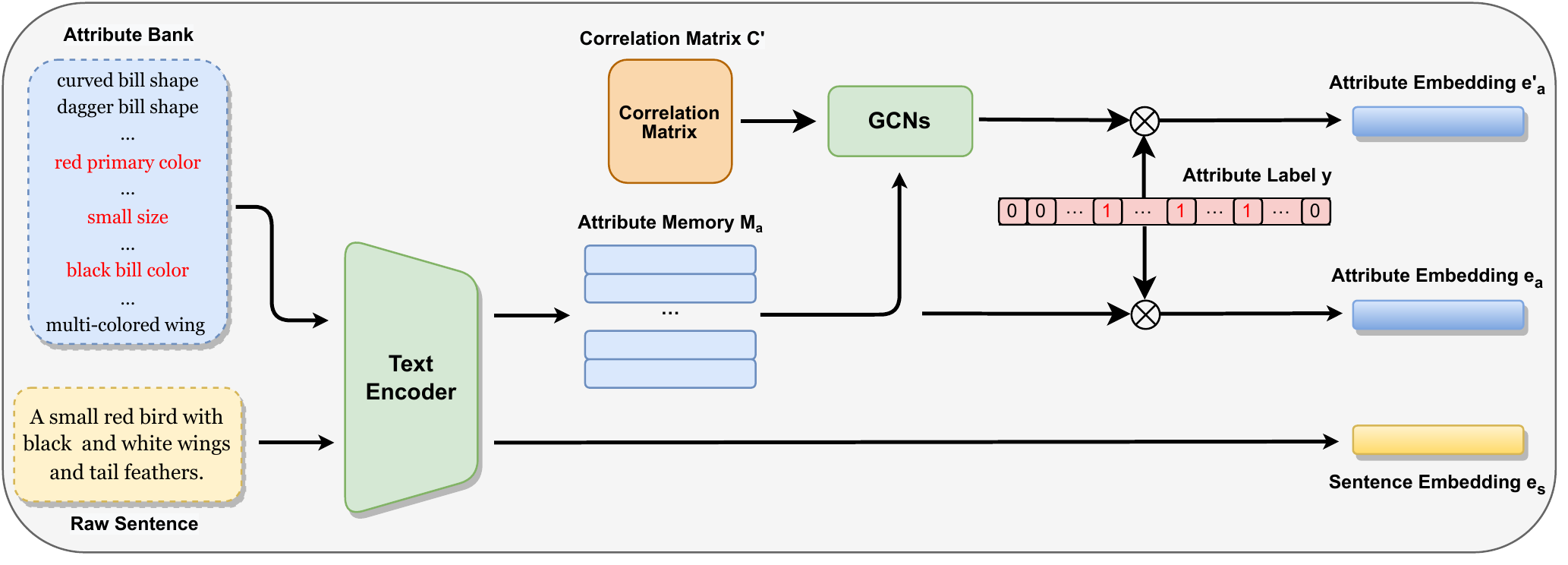}
	\caption{Illustration of attribute embedding construction.}
	\label{fig:attribute_embedding}
\end{figure*}

\begin{figure}[htbp]
	\centering
	\includegraphics[width=0.98\linewidth]{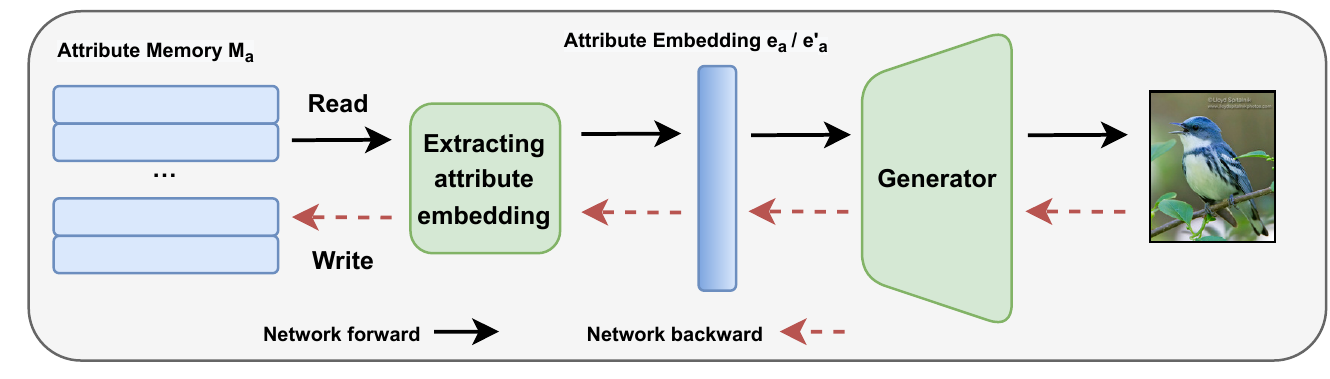}
	\caption{Illustration of the read-and-write process of the proposed memory.}
	\label{fig:memory_update}
\end{figure}

\subsection{Attribute Memory Construction}\label{sec:memory_construct}

In the previous text-to-image generation task, many works extract text embedding from only one sentence input, while ignore the attribute descriptions, which are critical factors for image-text matching. 
Therefore, we propose an attribute memory to assist the text-to-image generation task. 
In the following, we introduce the construction of the attribute memory and the read-and-write process of the proposed memory in detail.

We first collect all attribute descriptions and aggregate them together as an attribute bank. 
Then, we utilize a pre-trained text encoder~\cite{xu2018attngan} to extract all the descriptions as an attribute embedding memory $M_{a}$, in which each item represents the embedding of specific attribute description. Different from the previous memory modules~\cite{sukhbaatar2015end,zhao2021memory}, the read-and-write process of the proposed memory is achieved through forward inference and backward propagation of the generator, which is illustrated in~\cref{fig:memory_update}. 

Therefore, the attribute memory is first initialized by a set of attribute embeddings from the pre-trained text encoder, and then updated during the learning process. 
Several experiments are conducted to verify the importance of updated memory and attribute embedding initialization in~\cref{sec:ablation}.

\subsection{Attribute Memory Update}\label{sec:memory_update}

A fixed memory is difficult to be utilized for effective cross-modal image generation. Therefore, we design a learnable memory update scheme to update the attribute embedding memory dynamically in an effective way. In the following, we explore two different memory update mechanisms (sample-aware and sample-joint) to obtain suitable attribute embeddings from the attribute memory.

\textbf{Sample-aware memory update mechanism}:
In order to update the attribute memory, we treat all the parameters of the memory as the optimizable parameters and add them to the parameter group of the entire generator. Thus, the update of memory can be achieved through the gradient back-propagation of the network. 

As illustrated in~\cref{fig:attribute_embedding}, an image sample is annotated with a multi-attribute binary label $y$, in which $1$ denotes that the image has such an attribute while $0$ means it does not. Given a sample, $y$ is used to extract the corresponding embeddings from attribute memory $M_{a}$ and combine them together as a common attribute embedding $e_{a}$, which has the same dimension as the sentence embedding $e_{s}$. The construction of $e_{a} $ is denoted as: 
\begin{equation}\label{con:sample_aware}
	e_{a} = y\cdot M_{a},
\end{equation}
where $y\in \mathbb{R}^{1\times n}$, $M_{a}\in \mathbb{R}^{n\times d}$, $n$ represents the number of all attributes in a dataset, $d$ is the dimension of the embedding vector.
The combined attribute embedding $e_{a}$ is then set as an additional condition with sentence embedding $e_{s}$ together to jointly modulate the image features. 

In this way, through gradient back-propagation, only sample-related attribute embeddings are optimized when the network is updated. However, different attributes in the pre-defined attribute bank are inherently related. In fact, there exists an implicit knowledge graph between attributes beyond a single image. 
Combining the corresponding attributes with the way in \cref{con:sample_aware} only considers the co-occurrence of an image, while ignores the global correlation patterns within the entire dataset. In the following, we further propose a sample-joint memory update mechanism to model the global co-occurrence patterns. 




\textbf{Sample-joint memory update mechanism}:
Specifically, we introduce GCNs to model the relationship and propagate information between attributes based on the correlation matrix. In our GCN modeling, the attribute memory is set as the initial node features of the graph and each embedding denotes a node. Given the initial node features $H^{0}=M_{a}$ and the correlation matrix $C$, the GCN updates node features through stacked learnable transition matrix $W$. The forward of a GCN layer is denoted as:
\begin{equation}
	H^{l+1} = LeakyReLU(C\cdot H^{l}\cdot W^{l}),
\end{equation}
where $LeakyReLU(\cdot)$ is the Leaky ReLU activation function~\cite{xu2015empirical}, $W^{l}$ is the learnable parameter in the $l$-th layer of GCN, $H^{l}$ and $H^{l+1}$ are the input and output node features, respectively. 

In order to model the global correlation between attributes, we construct the correlation matrix $C$ by counting the occurrence of attribute pairs in the training set, which is denoted as:
\begin{equation}\label{con:binary_correlation_matrix}
	C_{ij}=\left\{
	\begin{aligned}
		0,\quad & {\rm {if}}\, P_{ij} < \tau \\
		1,\quad & {\rm {if}}\, P_{ij} \geq \tau 
	\end{aligned}\quad ,
	\right.
\end{equation}
where $P_{ij}$ represents the probability of $j$-th attribute when $i$-th appears in the whole dataset, $\tau$ is a threshold to filter noisy edges. As in \cite{chen2019multi}, we also apply the re-weighted scheme to alleviate the over-smoothing problem in binary correlation matrix. Thus, a re-weighted correlation matrix $C^{'}$ is denoted as:
\begin{equation}\label{con:reweighted_correlation_matrix}
	C^{'}_{ij}=\left\{
	\begin{aligned}
		& \frac{p}{\sum_{i=1,i\neq j}^{n} C_{ij}},\quad & {\rm {if}}\, i\neq j \\
		& 1-p,\quad & {\rm {if}}\, i=j 
	\end{aligned}\quad ,
	\right.
\end{equation}
where $p$ is a fixed weight assigned to a node itself and other correlated nodes to decide whether to consider neighbor information.

Therefore, initial node features $M_{a}$ are updated through stacked GCNs based on $C^{'}$. The node features of the last layer $H^{L}$ is also pooled into an attribute embedding, which is denoted as:
\begin{equation}\label{con:sample_joint}
	e^{'}_{a} = y\cdot H^{L}.
\end{equation}
Thus, when updating the attribute embedding of the current sample, the co-occurrence attribute embeddings of other samples will also be optimized, leading to a more effective attribute memory. 

\subsection{Attribute-Image Alignment}\label{sec:alignment}

With the introduction of attribute embedding, it is important to align the attribute-image during the cross-modal training. Therefore, we introduce a contrastive learning loss to align attribute and image embedding in a common space. 
Formally, we take the cosine similarity $\cos(\cdot, \cdot)$ as the distance metric:
\begin{equation}
	\cos(u,v) = \frac{u^{T}v}{\Vert u \Vert \cdot \Vert v \Vert}.
\end{equation} 
Thus, the contrastive loss takes a pair of input ($x^{i}, e^{j}_{a}$) and minimizes the embedding distance when they are from the same pair (\ie, $i=j$) but maximizes the distance otherwise (\ie, $i\neq j$). 
\begin{equation}\label{con:constrastive}
	\mathcal{L}_{cl}(u, v) = - \frac{1}{m}\sum_{i=1}^{m} \log \frac{\exp (\cos(u^{i}, v^{i})/\eta)}{\sum_{j=1}^{m} \exp (\cos(u^{i}, v^{j})/\eta)},
\end{equation}
\begin{align}
	\begin{split}
		\mathcal{L}_{attr\_real} & = \mathcal{L}_{cl}(D_{img}(x), e^{'}_{a}), \\
		\mathcal{L}_{attr\_fake} & = \mathcal{L}_{cl}(D_{img}(x_{f}), e^{'}_{a}),
	\end{split}
\end{align}
where $m$ is the mini-batch size of input samples, $u^{i}$ is the $i$-th sample in a mini-batch of $u$, and $v^{j}$ refers to the $j$-th one of $v$,
$\eta$ is a temperature hyper-parameter, $D_{img}(\cdot)$ is the extraction process of image embedding in the discriminator, and projects the input image into image embedding. 

Additionally, we apply contrastive learning to other modal pairs simultaneously, including image with sentence, fake image with real image of the same description. 

\begin{align}
	\begin{split}
		\mathcal{L}_{sent\_real} & = \mathcal{L}_{cl}(D_{img}(x), e_{s}), \\
		\mathcal{L}_{sent\_fake} & = \mathcal{L}_{cl}(D_{img}(x_{f}), e_{s}),
	\end{split}
\end{align}
\begin{equation}
	\mathcal{L}_{img} = \mathcal{L}_{cl}(D_{img}(x), D_{img}(x_{f})).
\end{equation}


\begin{figure*}[!t]
	\centering
	\includegraphics[width=\linewidth]{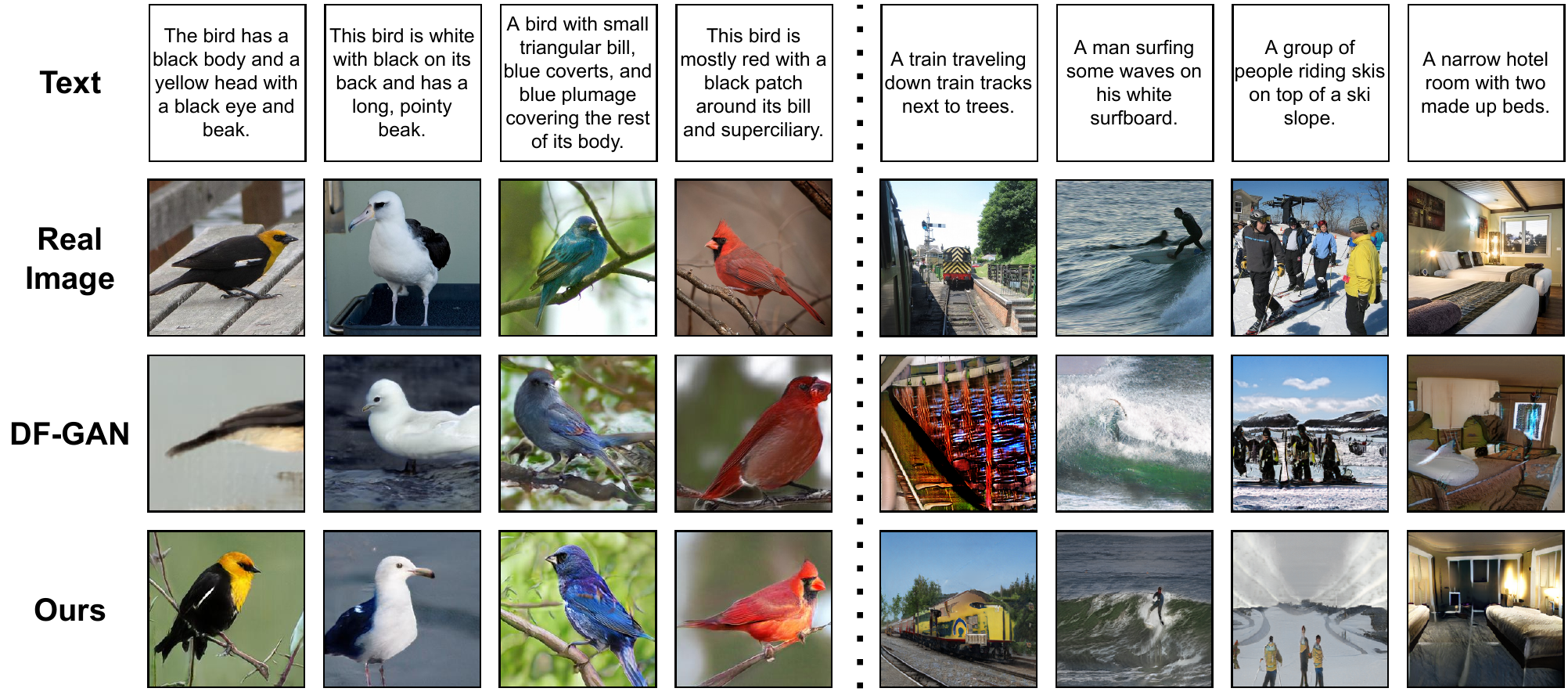}
	\caption{Qualitative comparisons between the proposed method and baseline DF-GAN.}
	\label{fig:qualitative_sota}
\end{figure*}

\subsection{Overall Optimization}\label{sec:optimization}

The proposed attribute-sentence-joint conditional framework is optimized in an adversarial manner that the generator and discriminator are asynchronously updated. We improve the ability of the proposed cGAN through three types of constraints: 1) authenticity discrimination; 2) multi-attribute classification; 3) cross-modal alignment.  All the loss functions are described in the following.

\textbf{Authenticity discrimination}: 
We employ the hinge loss~\cite{lim2017geometric} as the adversarial loss to stabilize the training process. 
The generator $G$ takes Gaussian noise $e_{z}$, attribute and sentence joint conditional embeddings $e_{s},e_{a}$ as input, 
while the discriminator $D$ is required to distinguish between $x$ and $x_{f}=G(e_{z},e_{s},e_{a})$. The corresponding adversarial loss functions for the generator $\mathcal{L}_{adv\_G}$ and discriminator $\mathcal{L}_{adv\_D}$ can be represented as follows:
\begin{equation}
	\begin{split}
		\mathcal{L}_{adv\_D} & = \mathbb{E}[\max(0, 1-D(x))] + \mathbb{E}[\max(0, 1+D(x_{f}))], \\
		\mathcal{L}_{adv\_G} & = - \mathbb{E}[D(x_{f})].
	\end{split}
\end{equation}

\textbf{Multi-attribute classification}:
The multi-attribute classification is set as an auxiliary task to let discriminator learn to recognize multiple attributes from a given image. 
In order to eliminate the bias in the learning process of two different tasks, we follow~\cite{hou2021conditional} to make the classifier capable of distinguishing real from fake while classifying attribute labels. The discriminative classifier maps $\mathcal{X}\rightarrow \mathcal{Y}\times \{0, 1\}$ ($n\times 2$ classes) that recognizes the attribute labels discriminatively. Specifically, when given real image $x$ (reps. fake image $x_{f}$), we convert it into logit $l_{r}$ (reps. $l_{f}$) through the discriminative classifier. Correspondingly, the label $y$ is extended twice that the odd positions denote fake label $y_{f}$ while the even ones refer to real label $y_{r}$. The above four items are $\in \mathbb{R}^{1\times 2n}$. 
We compute the BCE loss in generator $G$ and discriminator $D$ through:
\begin{equation}
	\mathcal{L}_{bce}(l,y) = - \frac{1}{2n}\sum_{i=1}^{2n}(y^{i}\log(l^{i}) + (1 - y^{i})\log(1 - l^{i})),
\end{equation}
\begin{equation}
	\begin{split}
		\mathcal{L}_{cls\_D} & = \mathcal{L}_{bce}(l_{r},y_{r}) + \mathcal{L}_{bce}(l_{f},y_{f}), \\
		\mathcal{L}_{cls\_G} & = \mathcal{L}_{bce}(l_{f},y_{r}) - \mathcal{L}_{bce}(l_{f},y_{f}),
	\end{split}
\end{equation}
where $n$ is the number of attributes, $y^{i}$ and $l^{i}$ are the $i$-th attribute of the attribute label and logit, respectively. $\mathcal{L}_{cls\_D}$ enables the discriminator to identify fake samples when doing multi-attribute classification, while $\mathcal{L}_{cls\_G}$ tries to fool D without identification.

\textbf{Cross-modal alignment}:
We combine all the contrastive loss functions between a real image $x$ and the corresponding text embeddings $\{e_{s}, e_{a}\}$ of $x$ to optimize the discriminator $D$. In addition, the contrastive loss functions between the fake image $x_{f}$ and $\{e_{s}, e_{a}\}$ are used to regularize the generator $G$. The corresponding alignment loss functions for $G$: $\mathcal{L}_{align\_G}$ and $D$: $\mathcal{L}_{align\_D}$ are denoted as:
\begin{equation}
	\begin{split}
		\mathcal{L}_{align\_D} & = \mathcal{L}_{attr\_real} + \mathcal{L}_{sent\_real}, \\
		\mathcal{L}_{align\_G} & = \mathcal{L}_{attr\_fake} + \mathcal{L}_{sent\_fake} + \mathcal{L}_{img}. 
	\end{split}
\end{equation}

Overall, the full objective function of the generator $\mathcal{L}_{G}$ and discriminator $\mathcal{L}_{D}$ are obtained as a weighted combination of the corresponding individual loss functions defined above. 
\begin{equation}\label{con:total}
	\begin{split}
		\mathcal{L}_{D} & = \mathcal{L}_{adv\_D} + \lambda_1 \mathcal{L}_{align\_D} + \lambda_2 \mathcal{L}_{cls\_D} + \lambda_3 \mathcal{L}_{ma-gp}, \\
		\mathcal{L}_{G} & = \mathcal{L}_{adv\_G} + \lambda_4 \mathcal{L}_{align\_G} + \lambda_5 \mathcal{L}_{cls\_G},
	\end{split}
\end{equation}
where $\lambda_1$, $\lambda_2$, $\lambda_3$, $\lambda_4$, $\lambda_5$ are the coefficient weights. $\mathcal{L}_{ma-gp}$ is the matching-aware gradient penalty in \cite{tao2020df}, which applies the gradient penalty on real images with the matching sentences.
\section{Experiment}


\subsection{Experimental Setup}\label{sec:setup}
In the following, we clearly introduce the datasets, evaluation metrics and implementation details.

\textbf{Datasets}:
To evaluate the capability of our method, we conduct extensive experiments on both Caltech-UCSD Birds 200 (CUB)~\cite{welinder2010caltech} and MS COCO~\cite{lin2014microsoft} datasets. The CUB dataset contains 200 bird categories with 312 attributes description. It is divided into training set and testing set with 8855 and 2933 images, respectively. Each image is annotated with 10 text captions and a multi-attribute binary label. As for the COCO, it is a multi-object dataset with 80 annotated categories. The entire dataset consists of $80k$ images for training and $40k$ images for testing. Each image is annotated with 5 captions and a multi-object label. 

\begin{figure}[!t]
	\centering
	\includegraphics[width=0.85\linewidth]{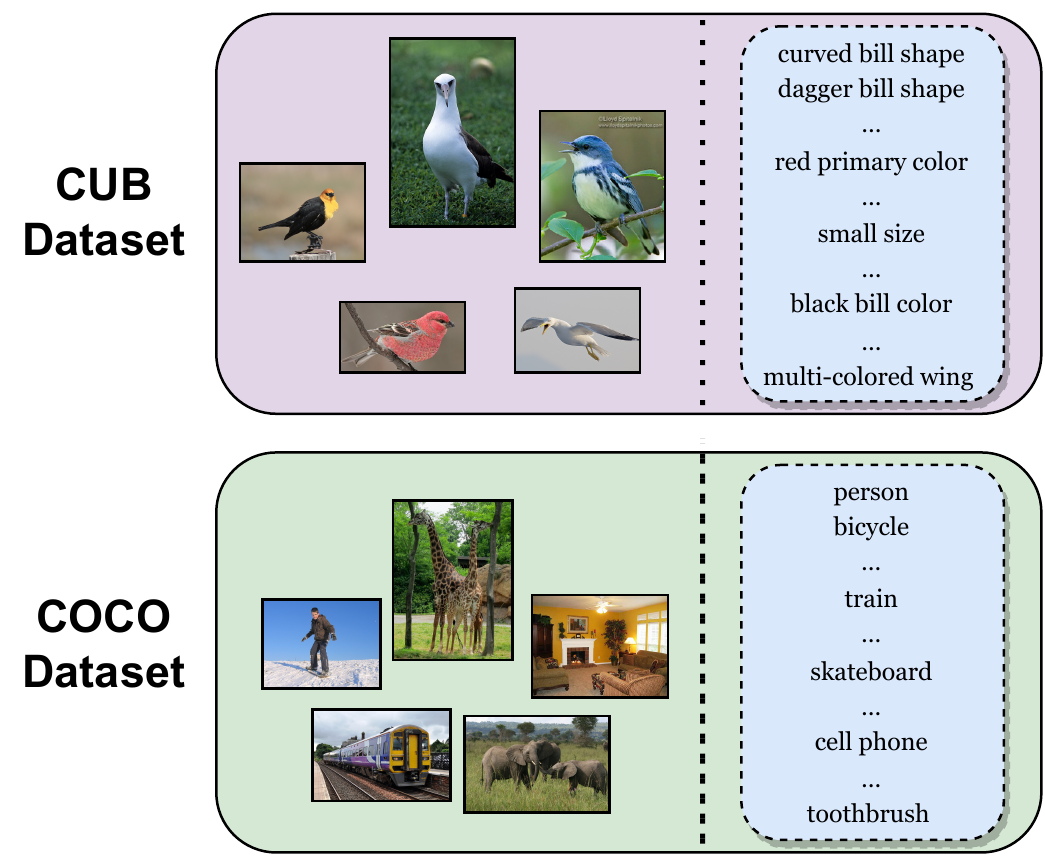}
	\caption{Datasets display. 
	Attribute definitions are different referring to CUB and COCO datasets.
	Left: image examples; Right: attribute descriptions.}
	\label{fig:dataset}
\end{figure}

\textbf{Evaluation Metrics}:
Following other text-to-image generation tasks~\cite{xu2018attngan,tao2020df}, we introduce Fréchet Inception Distance (FID)~\cite{heusel2017gans} and Inception Score (IS)~\cite{salimans2016improved} for image quality evaluation. 
Both of them utilize a pre-trained Inception-v3 network~\cite{szegedy2016rethinking} to extract the image features. 
Finally, in order to estimate the semantic consistency between the generated images and text inputs, we introduce two additional metrics. One is the top-1 image-sentence retrieval accuracy (Top-1 Acc) using the image and text encoder from~\cite{xu2018attngan}, and the other is the mean average precision (mAP) to evaluate the multi-attribute classification performance of generated images following~\cite{lanchantin2021general}.

\textbf{Implementation Details}:
We evaluate our methods on two widely used dataset, CUB and COCO. As shown in~\cref{fig:dataset}, CUB is a single-object dataset, in which each image contains a single bird with multiple part descriptions. We define attributes as these part descriptions such as red eyes, black wings \etc. In addition, COCO is a multi-object dataset, in which each image contains multiple objects. We define attributes as the instance object description, such as human, train \etc.
For each dataset, we first collect all the attribute descriptions and aggregate them together as an attribute bank, as shown in the right part of~\cref{fig:dataset}. Then, we utilize a pre-trained text encoder provided from AttnGAN~\cite{xu2018attngan} to embed the attribute descriptions and form an attribute memory. Next, the attribute embedding is obtained through~\cref{con:sample_aware} or~\cref{con:sample_joint}, and set as the condition with sentence embedding to jointly control the conditional text-to-image GAN.

We utilize DF-GAN~\cite{tao2020df} as the network backbone and design three effective strategies to improve the model performance. All the hyper-parameter settings are the same for both datasets, except for the total iteration steps. The model is optimized by the Adam optimizer~\cite{kingma2014adam} with exponential decay rates $(\beta_1=0.0, \beta_2=0.9)$. The learning rates are set to 0.0001 and 0.0004 for the generator and discriminator, respectively. According to~\cite{chen2019multi}, $\tau$ in~\cref{con:binary_correlation_matrix} is set to 0.4 and $p$ in~\cref{con:reweighted_correlation_matrix} is 0.25. In~\cref{con:constrastive}, $\eta$ is set to 0.1 as~\cite{zhang2021cross}. In~\cref{con:total}, the coefficient weights are set to $\lambda_1=\lambda_4=0.1,\lambda_2=\lambda_5=0.5,\lambda_3=2$.

\begin{table}[!t]
	\centering
	\caption{Comparison with State-of-the-art Methods on CUB and COCO datasets. The \textbf{black blod} results refer to the \textbf{best} performance.}
		\begin{tabular}{lccccc}
		\hline
  	\multirow{2}{*}{Method}  & \multirow{2}{*}{Reference} & \multicolumn{2}{c}{CUB} & \multicolumn{2}{c}{COCO} \\ 
  	\cline{3-6}
    &                        & FID $\downarrow$  & IS $\uparrow$ & FID $\downarrow$  & IS $\uparrow$ \\ 
  	\hline
		AttnGAN~\cite{xu2018attngan}   & CVPR18 & 23.98 & 4.36 & 35.49 & 25.89 \\
		DM-GAN~\cite{zhu2019dm}        & CVPR19 & 16.09 & 4.75 & 32.64 & 30.49 \\
		SD-GAN~\cite{yin2019semantics} & CVPR19 & -      & 4.67 & -     & 35.69 \\
		SEGAN~\cite{tan2019semantics}  & ICCV19 & 18.17 & 4.67 & 32.28 & 27.86 \\
		VICTR~\cite{han2020victr} & COLING20 & - & -    & 32.37 & 32.37 \\
		CPGAN~\cite{liang2020cpgan}    & ECCV20 &  -    & -    & 50.68 & \textbf{52.73} \\
		OP-GAN~\cite{hinz2019semantic} & TPAMI20 & -     & -    & 25.80 & 27.90 \\
		DAE-GAN~\cite{tao2020df}     & ICCV21 & 15.19 & 4.42 & 28.12 & 35.08 \\
		CL~\cite{ye2021improving} & BMVC21 & 14.38 & 4.77 & 20.79 & 33.34 \\
		MDD~\cite{feng2021modality} & TMM21 & 15.76 & 4.86 & 24.30 & 34.46 \\
		KD-GAN~\cite{peng2021knowledge} & TMM21  & 13.89 & 4.90 & 23.92 & 34.01 \\
		DF-GAN~\cite{tao2020df}      & CVPR22 & 14.81 & 5.10 & 21.42 & - \\
		\hline
		Ours                        &      & \textbf{8.57} & \textbf{5.28} & \textbf{12.39}  & 29.07 \\
		\bottomrule
	\end{tabular}
	\label{tab:sota}%
\end{table}%

\subsection{Comparison with SOTA}\label{sec:sota}

We compare our method with other previous text-to-image generation works in terms of both the qualitative and quantitative results.

\textbf{Qualitative Results}:
As shown in~\cref{fig:qualitative_sota}, the proposed method generates more semantically consistent images compared with the baseline DF-GAN. As for the CUB dataset, DF-GAN even synthesizes irregular birds body while our method avoids it well in column 1-2. As for the COCO dataset, we produce more reasonable images with better details in column 5-8. Visually, the images generated by our method are generally more realistic. For more visualizations, please refer to the supplementary material. 

\textbf{Quantitative Results}:
For fairness, we compare our method with previous state-of-the-art methods using the same pre-trained text encoder~\cite{xu2018attngan}, and the results are demonstrated in~\cref{tab:sota}. Our method achieves the best performance on CUB-FID, CUB-IS, and COCO-FID. Especially for the FID metric, the proposed method is much ahead of previous works, and degrades FID from 14.81 to 8.57 on the CUB dataset and from 21.42 to 12.39 on the COCO dataset. As for COCO-IS, our method achieves a comparable result. It is worth mentioning that COCO-IS utilizes a single-object pre-trained inception network to estimate a multi-object dataset, which will lead to inaccurate assessments~\cite{dinh2021tise}. Thus, relatively speaking, this does not objectively reflect the generated image quality on the COCO dataset. How to design a better multi-object image evaluation metric is a problem worth exploring.

\subsection{Ablation Study}\label{sec:ablation}

We conduct the ablation study to estimate the performance of each component in our method and demonstrate the experimental comparisons under different settings.


\textbf{Component Evaluation}:
The proposed method is composed of three main components, including the sample-aware and sample-joint mechanisms, and attribute-sentence-image alignment. The ablation study of each component is shown in~\cref{tab:component}, and the FID scores indicate the effectiveness of them compared to the baseline. The sample-joint strategy achieves better result than the sample-aware one since it models the global correlation and obtains a more suitable attribute memory. Furthermore, the combination of the sample-joint and alignment strategies gets the best result. 

\begin{table}[htbp]
	\centering
	\caption{Ablation Study on each proposed component. We compare the FID scores on each experiment list as follows. $\checkmark$ refers to the component to be use in our method. }
	\setlength{\tabcolsep}{1.6mm}{
	\begin{tabular}{c|c|c|c|c|c|c}
		\toprule
		Methods      & EXP0         & EXP1         & EXP2         & EXP3         & EXP4         & EXP5 \\
		\midrule
		Baseline     & $\checkmark$ & $\checkmark$ & $\checkmark$ & $\checkmark$ & $\checkmark$ & $\checkmark$ \\
		Sample-aware &              & $\checkmark$ &              &              & $\checkmark$ &              \\
		Sample-joint &              &              & $\checkmark$ &              &              & $\checkmark$ \\
		Alignment    &              &              &              & $\checkmark$ & $\checkmark$ & $\checkmark$ \\
		\midrule
		CUB-FID $\downarrow$ & 14.81 & 9.31        & 9.04         & 12.43        & 8.92       & \textbf{8.57}          \\
		\bottomrule
	\end{tabular}%
	}
	\label{tab:component}%
\end{table}%

\textbf{Importance of the attribute memory}:
We study the importance of the proposed attribute memory and reports the comparison results in~\cref{tab:importance_attr}. We utilize $n$ random noises to initialize the learnable memory in row 2. Row 3-4 represent the methods that use attribute embeddings as memory initialization. Row 3 uses a fixed memory while row 4 use a learnable one. 
According to the comparison results of row 2 and row 4, it is necessary to use attribute embeddings for memory initialization. Compared with row 3 and row 4, we found that using the updated memory instead of a fixed one facilitates model training and improves the performance well.

\begin{table}[htbp]
	\centering
	\caption{Evaluation of the importance of the proposed attribute memory using sample-aware update mechanism.}
	\setlength{\tabcolsep}{3.5mm}{
	\begin{tabular}{c|c}
		\toprule
		Settings                         & CUB-FID $\downarrow$ \\ 
		\midrule
		Baseline                         & 14.81  \\
		Random Initialized \& Updated    & 13.83  \\
		Attribute Initialized \& Fixed   & 11.81  \\ 
		Attribute Initialized \& Updated & \textbf{9.31} \\
		\bottomrule
	\end{tabular}%
	}
	\label{tab:importance_attr}%
\end{table}%

\textbf{Where to plug the sentence and attribute embeddings?}
We use attribute embedding to help improve the representation of a text input. Therefore, how to apply the sentence and attribute embedding to the network is a problem that needs to be explored. We try different combinations of these two embeddings and demonstrate the results in~\cref{tab:sent_attr}. The sentence embedding $e_{s}$ and attribute embedding $e_{a}$ are inserted into the up-fusion blocks, and each contains an up-sample operation with two affine transformation modules. The affine transformation manipulates visual feature maps conditioned on $e_{s}$ or $e_{a}$. Thus, we attempt to call one type (\aka, $OB-OT$) or two types (\aka, $OB-TT$) of embeddings in an up-fusion block. Overall, $OB-OT$ outperforms $OB-TT$, and in $OB-OT$, plugging the attribute embedding in lower layers while sentence embedding in higher layers gets the better result. 

\begin{table}[htbp]
	\centering
	\caption{Ablation Study on the combination of sentence and attribute embeddings. We compare the FID scores on each experiment list as follows. $OB-OT$ refers to one block contains one type of embedding, while $OB-TT$ denotes one block has two types of embedding.}
	\setlength{\tabcolsep}{3.5mm}{
	\begin{tabular}{c|c|c}
		\toprule
		\multicolumn{2}{c|}{Settings}                      & CUB-FID $\downarrow$ \\ 
		\midrule
		\multirow{2}{*}{OB-OT} & $e_{s}$-low, $e_{a}$-high & 9.67                          \\ 
		                       & $e_{a}$-low, $e_{s}$-high & \textbf{9.04}              \\
		\cline{1-2}
		\multirow{2}{*}{OB-TT} & $e_{s}$-front, $e_{a}$-behind & 9.63                    \\ 
		                       & $e_{a}$-front, $e_{s}$-behind & 9.94                 \\
		\bottomrule
	\end{tabular}%
	}
	\label{tab:sent_attr}%
\end{table}%

\begin{figure}[htbp]
	\centering
	\includegraphics[width=\linewidth]{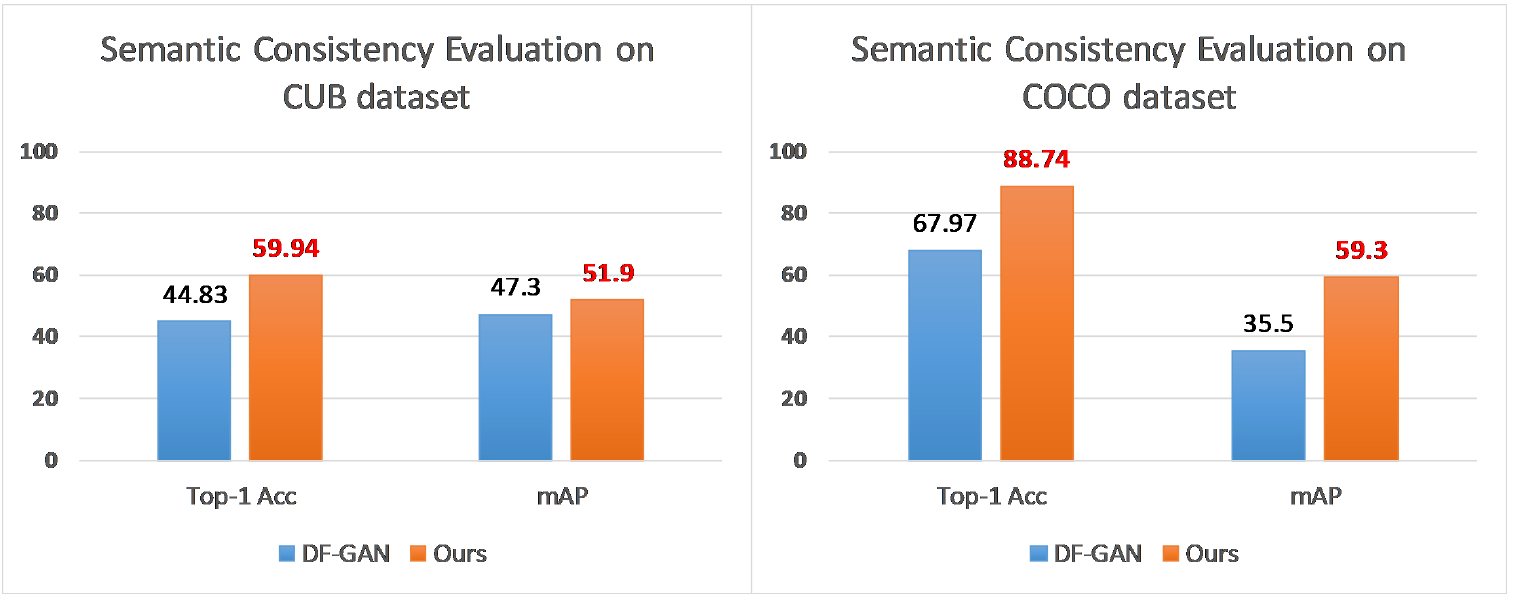}
	\caption{Semantic consistency performance evaluation of the proposed method and DF-GAN. Left: CUB dataset; Right: COCO dataset.}
	\label{fig:SCE}
\end{figure}

\textbf{Semantic consistency of cross modalities}:
Finally, we estimate the semantic consistency of the generated images between modalities, including sentence-image and attribute-image pairs. To measure the degree of matching between images and sentences input, we estimate the Top-1 image-text retrieval accuracy (Top-1 Acc) for both the proposed method and DF-GAN. As shown in~\cref{fig:SCE}, our method achieves higher Top-1 Acc than DF-GAN on both CUB and COCO datasets. This indicates that the generated images from our method match the given input sentences better. With the introduction of attribute representation, we also evaluate the multi-attribute classification performance of the generated images on relevant attributes using mAP. We train a multi-label classification network C-Tran~\cite{lanchantin2021general} as a multi-attribute classifier on both the CUB and COCO datasets. The compared results show that our method outperforms DF-GAN by a large margin, especially on COCO. This reveals that the generated images from our method have better attribute discriminability.


\subsection{Discussion}\label{sec:discussion}

We make a discussion about the interesting experimental findings, and analyze potential causes and improvements as follows.

\textbf{Multi-attribute conditional generation}:
In this paper, we introduce the attribute memory to augment text-to-image generation task and~\cref{tab:importance_attr} demonstrates that attribute memory plays an important role to improve the model performance. The generation thus is conditioned on sentence embedding and attribute embedding jointly.
Then, a question arises, what about the image is generated using only attribute embedding as a condition? To this end, we make an experiment under the sample-aware setting using attribute embedding only and achieve the FID score of 14.74, which is comparable to DF-GAN. It shows that this task is feasible to some extent. Multiple attribute descriptions provide the general content of a sample, and the sentence provides the association between attributes. The combination of the two synthesizes more photo-realistic and semantic matching images.

Similar to this setting, there are a few methods~\cite{bolluyt2019collapse,saseendran2021multi} studying multi-class conditional generation without sentence input. It takes a multi-class binary label as the condition input to generate the image that contains the given number of classes.
In fact, this task is more challenging due to the lack of association descriptions among multiple classes. 


\textbf{Sentence-only and sentence-with-attribute input}: The interface for practical application of the proposed method should be both sentence description and attribute label. However, in some cases, only sentence are provided. Surprisingly, our method is flexible for both the sentence-only and sentence-with-attribute input. Without additional labeling, the attribute label can be retrieved in the predefined attribute bank using the only sentence. In this paper, we use the sentence to receive top $k$ attribute descriptions according to the cosine similarity between the sentence embedding and each attribute embedding in the attribute bank. On the CUB dataset, we set $k$ to be 10, 30, 50, and obtain the FID score of 9.91, 8.89 and 9.11, respectively. Note that all these models are trained without contrastive learning. The model ($k$=10) trained with contrastive learning can even achieve an FID score of 8.78, which is comparable with the one trained with sentence-with-attribute input (FID=8.57). The experimental results indicate that our method also works well using only sentence as input, and is still superior to other compared methods.

\section{Conclusion}

In this paper, we have proposed Adma-GAN, an attribute-driven memory augmented GAN for the text-to-image generation task. It is able to synthesize photo-realistic and semantically consistent images. The main contribution is that we propose an effective text representation method with the complements of attribute information to assist in controlling image generation. Firstly, we construct an attribute memory to jointly control the text-to-image generation with sentence input. With the help of attribute memory, the input text representation is enriched, and the cross-modal gap is thus reduced. Secondly, we explore two memory update mechanisms, sample-aware and sample-joint mechanisms, to dynamically optimize a generalized attribute memory. The sample-joint mechanism outperforms the sample-aware one since it models the global correlation between attributes within a dataset. Thirdly, we employ the contrastive learning in attribute-to-image, sentence-to-image and image-to-image, to facilitate the cross-modal alignment. Combining all the above strategies, our method achieves substantial performance improvements on both the CUB and COCO datasets. 

\begin{acks}
  This work is supported in part by National Key Research and Development Program of China under Grant 2020AAA0107400, Zhejiang Provincial Natural Science Foundation of China under Grant LR19F020004, National Natural Science Foundation of China under Grant U20A20222.
\end{acks}

\bibliographystyle{ACM-Reference-Format}
\bibliography{reference}


\begin{thebibliography}{62}


\ifx \showCODEN    \undefined \def \showCODEN     #1{\unskip}     \fi
\ifx \showDOI      \undefined \def \showDOI       #1{#1}\fi
\ifx \showISBNx    \undefined \def \showISBNx     #1{\unskip}     \fi
\ifx \showISBNxiii \undefined \def \showISBNxiii  #1{\unskip}     \fi
\ifx \showISSN     \undefined \def \showISSN      #1{\unskip}     \fi
\ifx \showLCCN     \undefined \def \showLCCN      #1{\unskip}     \fi
\ifx \shownote     \undefined \def \shownote      #1{#1}          \fi
\ifx \showarticletitle \undefined \def \showarticletitle #1{#1}   \fi
\ifx \showURL      \undefined \def \showURL       {\relax}        \fi
\providecommand\bibfield[2]{#2}
\providecommand\bibinfo[2]{#2}
\providecommand\natexlab[1]{#1}
\providecommand\showeprint[2][]{arXiv:#2}

\bibitem[\protect\citeauthoryear{Antol, Agrawal, Lu, Mitchell, Batra, Zitnick,
  and Parikh}{Antol et~al\mbox{.}}{2015}]%
        {antol2015vqa}
\bibfield{author}{\bibinfo{person}{Stanislaw Antol}, \bibinfo{person}{Aishwarya
  Agrawal}, \bibinfo{person}{Jiasen Lu}, \bibinfo{person}{Margaret Mitchell},
  \bibinfo{person}{Dhruv Batra}, \bibinfo{person}{C~Lawrence Zitnick}, {and}
  \bibinfo{person}{Devi Parikh}.} \bibinfo{year}{2015}\natexlab{}.
\newblock \showarticletitle{Vqa: Visual question answering}. In
  \bibinfo{booktitle}{\emph{Proceedings of the IEEE international conference on
  computer vision}}. \bibinfo{pages}{2425--2433}.
\newblock


\bibitem[\protect\citeauthoryear{Bolluyt and Comaniciu}{Bolluyt and
  Comaniciu}{2019}]%
        {bolluyt2019collapse}
\bibfield{author}{\bibinfo{person}{Elijah Bolluyt} {and}
  \bibinfo{person}{Cristina Comaniciu}.} \bibinfo{year}{2019}\natexlab{}.
\newblock \showarticletitle{Collapse Resistant Deep Convolutional GAN for
  Multi-Object Image Generation}. In \bibinfo{booktitle}{\emph{2019 18th IEEE
  International Conference On Machine Learning And Applications (ICMLA)}}.
  IEEE, \bibinfo{pages}{1404--1408}.
\newblock


\bibitem[\protect\citeauthoryear{Cai, Wang, Yu, Li, Xu, Li, and Li}{Cai
  et~al\mbox{.}}{2019}]%
        {cai2019dualattn}
\bibfield{author}{\bibinfo{person}{Yali Cai}, \bibinfo{person}{Xiaoru Wang},
  \bibinfo{person}{Zhihong Yu}, \bibinfo{person}{Fu Li},
  \bibinfo{person}{Peirong Xu}, \bibinfo{person}{Yueli Li}, {and}
  \bibinfo{person}{Lixian Li}.} \bibinfo{year}{2019}\natexlab{}.
\newblock \showarticletitle{Dualattn-GAN: Text to image synthesis with dual
  attentional generative adversarial network}.
\newblock \bibinfo{journal}{\emph{IEEE Access}}  \bibinfo{volume}{7}
  (\bibinfo{year}{2019}), \bibinfo{pages}{183706--183716}.
\newblock


\bibitem[\protect\citeauthoryear{Chen, Liao, Chuang, Hsu, Fu, and Sun}{Chen
  et~al\mbox{.}}{2017}]%
        {chen2017show}
\bibfield{author}{\bibinfo{person}{Tseng-Hung Chen}, \bibinfo{person}{Yuan-Hong
  Liao}, \bibinfo{person}{Ching-Yao Chuang}, \bibinfo{person}{Wan-Ting Hsu},
  \bibinfo{person}{Jianlong Fu}, {and} \bibinfo{person}{Min Sun}.}
  \bibinfo{year}{2017}\natexlab{}.
\newblock \showarticletitle{Show, adapt and tell: Adversarial training of
  cross-domain image captioner}. In \bibinfo{booktitle}{\emph{Proceedings of
  the IEEE international conference on computer vision}}.
  \bibinfo{pages}{521--530}.
\newblock


\bibitem[\protect\citeauthoryear{Chen, Wei, Wang, and Guo}{Chen
  et~al\mbox{.}}{2019}]%
        {chen2019multi}
\bibfield{author}{\bibinfo{person}{Zhao-Min Chen}, \bibinfo{person}{Xiu-Shen
  Wei}, \bibinfo{person}{Peng Wang}, {and} \bibinfo{person}{Yanwen Guo}.}
  \bibinfo{year}{2019}\natexlab{}.
\newblock \showarticletitle{Multi-label image recognition with graph
  convolutional networks}. In \bibinfo{booktitle}{\emph{Proceedings of the
  IEEE/CVF conference on computer vision and pattern recognition}}.
  \bibinfo{pages}{5177--5186}.
\newblock


\bibitem[\protect\citeauthoryear{Cheng, Wu, Tian, Wang, and Tao}{Cheng
  et~al\mbox{.}}{2020}]%
        {cheng2020rifegan}
\bibfield{author}{\bibinfo{person}{Jun Cheng}, \bibinfo{person}{Fuxiang Wu},
  \bibinfo{person}{Yanling Tian}, \bibinfo{person}{Lei Wang}, {and}
  \bibinfo{person}{Dapeng Tao}.} \bibinfo{year}{2020}\natexlab{}.
\newblock \showarticletitle{RiFeGAN: Rich feature generation for text-to-image
  synthesis from prior knowledge}. In \bibinfo{booktitle}{\emph{Proceedings of
  the IEEE/CVF Conference on Computer Vision and Pattern Recognition}}.
  \bibinfo{pages}{10911--10920}.
\newblock


\bibitem[\protect\citeauthoryear{Ding, Yang, Hong, Zheng, Zhou, Yin, Lin, Zou,
  Shao, Yang, et~al\mbox{.}}{Ding et~al\mbox{.}}{2021}]%
        {ding2021cogview}
\bibfield{author}{\bibinfo{person}{Ming Ding}, \bibinfo{person}{Zhuoyi Yang},
  \bibinfo{person}{Wenyi Hong}, \bibinfo{person}{Wendi Zheng},
  \bibinfo{person}{Chang Zhou}, \bibinfo{person}{Da Yin},
  \bibinfo{person}{Junyang Lin}, \bibinfo{person}{Xu Zou},
  \bibinfo{person}{Zhou Shao}, \bibinfo{person}{Hongxia Yang}, {et~al\mbox{.}}}
  \bibinfo{year}{2021}\natexlab{}.
\newblock \showarticletitle{Cogview: Mastering text-to-image generation via
  transformers}.
\newblock \bibinfo{journal}{\emph{Advances in Neural Information Processing
  Systems}}  \bibinfo{volume}{34} (\bibinfo{year}{2021}).
\newblock


\bibitem[\protect\citeauthoryear{Dinh, Nguyen, and Hua}{Dinh
  et~al\mbox{.}}{2021}]%
        {dinh2021tise}
\bibfield{author}{\bibinfo{person}{Tan~M Dinh}, \bibinfo{person}{Rang Nguyen},
  {and} \bibinfo{person}{Binh-Son Hua}.} \bibinfo{year}{2021}\natexlab{}.
\newblock \showarticletitle{TISE: A Toolbox for Text-to-Image Synthesis
  Evaluation}.
\newblock \bibinfo{journal}{\emph{arXiv preprint arXiv:2112.01398}}
  (\bibinfo{year}{2021}).
\newblock


\bibitem[\protect\citeauthoryear{El, Licht, and Yosephian}{El
  et~al\mbox{.}}{2019}]%
        {el2019gilt}
\bibfield{author}{\bibinfo{person}{Ori~Bar El}, \bibinfo{person}{Ori Licht},
  {and} \bibinfo{person}{Netanel Yosephian}.} \bibinfo{year}{2019}\natexlab{}.
\newblock \showarticletitle{Gilt: Generating images from long text}.
\newblock \bibinfo{journal}{\emph{arXiv preprint arXiv:1901.02404}}
  (\bibinfo{year}{2019}).
\newblock


\bibitem[\protect\citeauthoryear{Feng, Niu, Li, and Wang}{Feng
  et~al\mbox{.}}{2021}]%
        {feng2021modality}
\bibfield{author}{\bibinfo{person}{Fangxiang Feng}, \bibinfo{person}{Tianrui
  Niu}, \bibinfo{person}{Ruifan Li}, {and} \bibinfo{person}{Xiaojie Wang}.}
  \bibinfo{year}{2021}\natexlab{}.
\newblock \showarticletitle{Modality Disentangled Discriminator for
  Text-to-Image Synthesis}.
\newblock \bibinfo{journal}{\emph{IEEE Transactions on Multimedia}}
  (\bibinfo{year}{2021}).
\newblock


\bibitem[\protect\citeauthoryear{Frolov, Jolly, Hees, and Dengel}{Frolov
  et~al\mbox{.}}{2020}]%
        {frolov2020leveraging}
\bibfield{author}{\bibinfo{person}{Stanislav Frolov}, \bibinfo{person}{Shailza
  Jolly}, \bibinfo{person}{J{\"o}rn Hees}, {and} \bibinfo{person}{Andreas
  Dengel}.} \bibinfo{year}{2020}\natexlab{}.
\newblock \showarticletitle{Leveraging visual question answering to improve
  text-to-image synthesis}. In \bibinfo{booktitle}{\emph{Proceedings of the
  Second Workshop on Beyond Vision and LANguage: inTEgrating Real-world
  kNowledge (LANTERN)}}. \bibinfo{pages}{17--22}.
\newblock


\bibitem[\protect\citeauthoryear{Gou, Wu, Li, Gong, and Han}{Gou
  et~al\mbox{.}}{2020}]%
        {gou2020segattngan}
\bibfield{author}{\bibinfo{person}{Yuchuan Gou}, \bibinfo{person}{Qiancheng
  Wu}, \bibinfo{person}{Minghao Li}, \bibinfo{person}{Bo Gong}, {and}
  \bibinfo{person}{Mei Han}.} \bibinfo{year}{2020}\natexlab{}.
\newblock \showarticletitle{SegAttnGAN: Text to image generation with
  segmentation attention}.
\newblock \bibinfo{journal}{\emph{arXiv preprint arXiv:2005.12444}}
  (\bibinfo{year}{2020}).
\newblock


\bibitem[\protect\citeauthoryear{Han, Long, Luo, Wang, and Poon}{Han
  et~al\mbox{.}}{2020}]%
        {han2020victr}
\bibfield{author}{\bibinfo{person}{Soyeon~Caren Han}, \bibinfo{person}{Siqu
  Long}, \bibinfo{person}{Siwen Luo}, \bibinfo{person}{Kunze Wang}, {and}
  \bibinfo{person}{Josiah Poon}.} \bibinfo{year}{2020}\natexlab{}.
\newblock \showarticletitle{VICTR: Visual Information Captured Text
  Representation for Text-to-Image Multimodal Tasks}.
\newblock \bibinfo{journal}{\emph{arXiv preprint arXiv:2010.03182}}
  (\bibinfo{year}{2020}).
\newblock


\bibitem[\protect\citeauthoryear{He, Zhang, Ren, and Sun}{He
  et~al\mbox{.}}{2016}]%
        {he2016deep}
\bibfield{author}{\bibinfo{person}{Kaiming He}, \bibinfo{person}{Xiangyu
  Zhang}, \bibinfo{person}{Shaoqing Ren}, {and} \bibinfo{person}{Jian Sun}.}
  \bibinfo{year}{2016}\natexlab{}.
\newblock \showarticletitle{Deep residual learning for image recognition}. In
  \bibinfo{booktitle}{\emph{Proceedings of the IEEE conference on computer
  vision and pattern recognition}}. \bibinfo{pages}{770--778}.
\newblock


\bibitem[\protect\citeauthoryear{Heusel, Ramsauer, Unterthiner, Nessler, and
  Hochreiter}{Heusel et~al\mbox{.}}{2017}]%
        {heusel2017gans}
\bibfield{author}{\bibinfo{person}{Martin Heusel}, \bibinfo{person}{Hubert
  Ramsauer}, \bibinfo{person}{Thomas Unterthiner}, \bibinfo{person}{Bernhard
  Nessler}, {and} \bibinfo{person}{Sepp Hochreiter}.}
  \bibinfo{year}{2017}\natexlab{}.
\newblock \showarticletitle{Gans trained by a two time-scale update rule
  converge to a local nash equilibrium}.
\newblock \bibinfo{journal}{\emph{Advances in neural information processing
  systems}}  \bibinfo{volume}{30} (\bibinfo{year}{2017}).
\newblock


\bibitem[\protect\citeauthoryear{Hinz, Heinrich, and Wermter}{Hinz
  et~al\mbox{.}}{2019}]%
        {hinz2019semantic}
\bibfield{author}{\bibinfo{person}{Tobias Hinz}, \bibinfo{person}{Stefan
  Heinrich}, {and} \bibinfo{person}{Stefan Wermter}.}
  \bibinfo{year}{2019}\natexlab{}.
\newblock \showarticletitle{Semantic object accuracy for generative
  text-to-image synthesis}.
\newblock \bibinfo{journal}{\emph{arXiv preprint arXiv:1910.13321}}
  (\bibinfo{year}{2019}).
\newblock


\bibitem[\protect\citeauthoryear{Hong, Yang, Choi, and Lee}{Hong
  et~al\mbox{.}}{2018}]%
        {hong2018inferring}
\bibfield{author}{\bibinfo{person}{Seunghoon Hong}, \bibinfo{person}{Dingdong
  Yang}, \bibinfo{person}{Jongwook Choi}, {and} \bibinfo{person}{Honglak Lee}.}
  \bibinfo{year}{2018}\natexlab{}.
\newblock \showarticletitle{Inferring semantic layout for hierarchical
  text-to-image synthesis}. In \bibinfo{booktitle}{\emph{Proceedings of the
  IEEE conference on computer vision and pattern recognition}}.
  \bibinfo{pages}{7986--7994}.
\newblock


\bibitem[\protect\citeauthoryear{Hou, Cao, Shen, and Cheng}{Hou
  et~al\mbox{.}}{2021}]%
        {hou2021conditional}
\bibfield{author}{\bibinfo{person}{Liang Hou}, \bibinfo{person}{Qi Cao},
  \bibinfo{person}{Huawei Shen}, {and} \bibinfo{person}{Xueqi Cheng}.}
  \bibinfo{year}{2021}\natexlab{}.
\newblock \showarticletitle{Conditional GANs with Auxiliary Discriminative
  Classifier}.
\newblock \bibinfo{journal}{\emph{arXiv preprint arXiv:2107.10060}}
  (\bibinfo{year}{2021}).
\newblock


\bibitem[\protect\citeauthoryear{Kim, Joo, and Kim}{Kim et~al\mbox{.}}{2020}]%
        {kim2020tivgan}
\bibfield{author}{\bibinfo{person}{Doyeon Kim}, \bibinfo{person}{Donggyu Joo},
  {and} \bibinfo{person}{Junmo Kim}.} \bibinfo{year}{2020}\natexlab{}.
\newblock \showarticletitle{TiVGAN: Text to Image to Video Generation With
  Step-by-Step Evolutionary Generator}.
\newblock \bibinfo{journal}{\emph{IEEE Access}}  \bibinfo{volume}{8}
  (\bibinfo{year}{2020}), \bibinfo{pages}{153113--153122}.
\newblock


\bibitem[\protect\citeauthoryear{Kingma and Ba}{Kingma and Ba}{2014}]%
        {kingma2014adam}
\bibfield{author}{\bibinfo{person}{Diederik~P Kingma} {and}
  \bibinfo{person}{Jimmy Ba}.} \bibinfo{year}{2014}\natexlab{}.
\newblock \showarticletitle{Adam: A method for stochastic optimization}.
\newblock \bibinfo{journal}{\emph{arXiv preprint arXiv:1412.6980}}
  (\bibinfo{year}{2014}).
\newblock


\bibitem[\protect\citeauthoryear{Kipf and Welling}{Kipf and Welling}{2016}]%
        {kipf2016semi}
\bibfield{author}{\bibinfo{person}{Thomas~N Kipf} {and} \bibinfo{person}{Max
  Welling}.} \bibinfo{year}{2016}\natexlab{}.
\newblock \showarticletitle{Semi-supervised classification with graph
  convolutional networks}.
\newblock \bibinfo{journal}{\emph{arXiv preprint arXiv:1609.02907}}
  (\bibinfo{year}{2016}).
\newblock


\bibitem[\protect\citeauthoryear{Lanchantin, Wang, Ordonez, and Qi}{Lanchantin
  et~al\mbox{.}}{2021}]%
        {lanchantin2021general}
\bibfield{author}{\bibinfo{person}{Jack Lanchantin}, \bibinfo{person}{Tianlu
  Wang}, \bibinfo{person}{Vicente Ordonez}, {and} \bibinfo{person}{Yanjun Qi}.}
  \bibinfo{year}{2021}\natexlab{}.
\newblock \showarticletitle{General multi-label image classification with
  transformers}. In \bibinfo{booktitle}{\emph{Proceedings of the IEEE/CVF
  Conference on Computer Vision and Pattern Recognition}}.
  \bibinfo{pages}{16478--16488}.
\newblock


\bibitem[\protect\citeauthoryear{Li, Torr, and Lukasiewicz}{Li
  et~al\mbox{.}}{2021}]%
        {li2021memory}
\bibfield{author}{\bibinfo{person}{Bowen Li}, \bibinfo{person}{Philip Torr},
  {and} \bibinfo{person}{Thomas Lukasiewicz}.} \bibinfo{year}{2021}\natexlab{}.
\newblock \showarticletitle{Memory-Driven Text-to-Image Generation}.
\newblock  (\bibinfo{year}{2021}).
\newblock


\bibitem[\protect\citeauthoryear{Li, Zhang, Zhang, Huang, He, Lyu, and Gao}{Li
  et~al\mbox{.}}{2019}]%
        {li2019object}
\bibfield{author}{\bibinfo{person}{Wenbo Li}, \bibinfo{person}{Pengchuan
  Zhang}, \bibinfo{person}{Lei Zhang}, \bibinfo{person}{Qiuyuan Huang},
  \bibinfo{person}{Xiaodong He}, \bibinfo{person}{Siwei Lyu}, {and}
  \bibinfo{person}{Jianfeng Gao}.} \bibinfo{year}{2019}\natexlab{}.
\newblock \showarticletitle{Object-driven text-to-image synthesis via
  adversarial training}. In \bibinfo{booktitle}{\emph{Proceedings of the
  IEEE/CVF Conference on Computer Vision and Pattern Recognition}}.
  \bibinfo{pages}{12174--12182}.
\newblock


\bibitem[\protect\citeauthoryear{Liang, Pei, and Lu}{Liang
  et~al\mbox{.}}{2020}]%
        {liang2020cpgan}
\bibfield{author}{\bibinfo{person}{Jiadong Liang}, \bibinfo{person}{Wenjie
  Pei}, {and} \bibinfo{person}{Feng Lu}.} \bibinfo{year}{2020}\natexlab{}.
\newblock \showarticletitle{Cpgan: Content-parsing generative adversarial
  networks for text-to-image synthesis}. In \bibinfo{booktitle}{\emph{European
  Conference on Computer Vision}}. Springer, \bibinfo{pages}{491--508}.
\newblock


\bibitem[\protect\citeauthoryear{Lim and Ye}{Lim and Ye}{2017}]%
        {lim2017geometric}
\bibfield{author}{\bibinfo{person}{Jae~Hyun Lim} {and}
  \bibinfo{person}{Jong~Chul Ye}.} \bibinfo{year}{2017}\natexlab{}.
\newblock \showarticletitle{Geometric gan}.
\newblock \bibinfo{journal}{\emph{arXiv preprint arXiv:1705.02894}}
  (\bibinfo{year}{2017}).
\newblock


\bibitem[\protect\citeauthoryear{Lin, Maire, Belongie, Hays, Perona, Ramanan,
  Doll{\'a}r, and Zitnick}{Lin et~al\mbox{.}}{2014}]%
        {lin2014microsoft}
\bibfield{author}{\bibinfo{person}{Tsung-Yi Lin}, \bibinfo{person}{Michael
  Maire}, \bibinfo{person}{Serge Belongie}, \bibinfo{person}{James Hays},
  \bibinfo{person}{Pietro Perona}, \bibinfo{person}{Deva Ramanan},
  \bibinfo{person}{Piotr Doll{\'a}r}, {and} \bibinfo{person}{C~Lawrence
  Zitnick}.} \bibinfo{year}{2014}\natexlab{}.
\newblock \showarticletitle{Microsoft coco: Common objects in context}. In
  \bibinfo{booktitle}{\emph{European conference on computer vision}}. Springer,
  \bibinfo{pages}{740--755}.
\newblock


\bibitem[\protect\citeauthoryear{Mirza and Osindero}{Mirza and
  Osindero}{2014}]%
        {mirza2014conditional}
\bibfield{author}{\bibinfo{person}{Mehdi Mirza} {and} \bibinfo{person}{Simon
  Osindero}.} \bibinfo{year}{2014}\natexlab{}.
\newblock \showarticletitle{Conditional generative adversarial nets}.
\newblock \bibinfo{journal}{\emph{arXiv preprint arXiv:1411.1784}}
  (\bibinfo{year}{2014}).
\newblock


\bibitem[\protect\citeauthoryear{NVIDIA}{NVIDIA}{2021}]%
        {nvidia2021gaugan2}
\bibfield{author}{\bibinfo{person}{NVIDIA}.} \bibinfo{year}{2021}\natexlab{}.
\newblock \bibinfo{booktitle}{\emph{Gaugan2}}.
\newblock
\urldef\tempurl%
\url{http://gaugan.org/gaugan2/}
\showURL{%
\tempurl}


\bibitem[\protect\citeauthoryear{Peng, Zhou, Sun, Cao, Wu, Huang, and Ji}{Peng
  et~al\mbox{.}}{2021}]%
        {peng2021knowledge}
\bibfield{author}{\bibinfo{person}{Jun Peng}, \bibinfo{person}{Yiyi Zhou},
  \bibinfo{person}{Xiaoshuai Sun}, \bibinfo{person}{Liujuan Cao},
  \bibinfo{person}{Yongjian Wu}, \bibinfo{person}{Feiyue Huang}, {and}
  \bibinfo{person}{Rongrong Ji}.} \bibinfo{year}{2021}\natexlab{}.
\newblock \showarticletitle{Knowledge-Driven Generative Adversarial Network for
  Text-to-Image Synthesis}.
\newblock \bibinfo{journal}{\emph{IEEE Transactions on Multimedia}}
  (\bibinfo{year}{2021}).
\newblock


\bibitem[\protect\citeauthoryear{Qiao, Zhang, Xu, and Tao}{Qiao
  et~al\mbox{.}}{2019}]%
        {qiao2019mirrorgan}
\bibfield{author}{\bibinfo{person}{Tingting Qiao}, \bibinfo{person}{Jing
  Zhang}, \bibinfo{person}{Duanqing Xu}, {and} \bibinfo{person}{Dacheng Tao}.}
  \bibinfo{year}{2019}\natexlab{}.
\newblock \showarticletitle{Mirrorgan: Learning text-to-image generation by
  redescription}. In \bibinfo{booktitle}{\emph{Proceedings of the IEEE/CVF
  Conference on Computer Vision and Pattern Recognition}}.
  \bibinfo{pages}{1505--1514}.
\newblock


\bibitem[\protect\citeauthoryear{Ramesh, Pavlov, Goh, Gray, Voss, Radford,
  Chen, and Sutskever}{Ramesh et~al\mbox{.}}{2021}]%
        {ramesh2021zero}
\bibfield{author}{\bibinfo{person}{Aditya Ramesh}, \bibinfo{person}{Mikhail
  Pavlov}, \bibinfo{person}{Gabriel Goh}, \bibinfo{person}{Scott Gray},
  \bibinfo{person}{Chelsea Voss}, \bibinfo{person}{Alec Radford},
  \bibinfo{person}{Mark Chen}, {and} \bibinfo{person}{Ilya Sutskever}.}
  \bibinfo{year}{2021}\natexlab{}.
\newblock \showarticletitle{Zero-shot text-to-image generation}. In
  \bibinfo{booktitle}{\emph{International Conference on Machine Learning}}.
  PMLR, \bibinfo{pages}{8821--8831}.
\newblock
\urldef\tempurl%
\url{https://openai.com/blog/dall-e/}
\showURL{%
\tempurl}


\bibitem[\protect\citeauthoryear{Ruan, Zhang, Zhang, Fan, Tang, Liu, and
  Chen}{Ruan et~al\mbox{.}}{2021}]%
        {ruan2021dae}
\bibfield{author}{\bibinfo{person}{Shulan Ruan}, \bibinfo{person}{Yong Zhang},
  \bibinfo{person}{Kun Zhang}, \bibinfo{person}{Yanbo Fan},
  \bibinfo{person}{Fan Tang}, \bibinfo{person}{Qi Liu}, {and}
  \bibinfo{person}{Enhong Chen}.} \bibinfo{year}{2021}\natexlab{}.
\newblock \showarticletitle{Dae-gan: Dynamic aspect-aware gan for text-to-image
  synthesis}. In \bibinfo{booktitle}{\emph{Proceedings of the IEEE/CVF
  International Conference on Computer Vision}}. \bibinfo{pages}{13960--13969}.
\newblock


\bibitem[\protect\citeauthoryear{Salimans, Goodfellow, Zaremba, Cheung,
  Radford, and Chen}{Salimans et~al\mbox{.}}{2016}]%
        {salimans2016improved}
\bibfield{author}{\bibinfo{person}{Tim Salimans}, \bibinfo{person}{Ian
  Goodfellow}, \bibinfo{person}{Wojciech Zaremba}, \bibinfo{person}{Vicki
  Cheung}, \bibinfo{person}{Alec Radford}, {and} \bibinfo{person}{Xi Chen}.}
  \bibinfo{year}{2016}\natexlab{}.
\newblock \showarticletitle{Improved techniques for training gans}.
\newblock \bibinfo{journal}{\emph{Advances in neural information processing
  systems}}  \bibinfo{volume}{29} (\bibinfo{year}{2016}).
\newblock


\bibitem[\protect\citeauthoryear{Salvador, Hynes, Aytar, Marin, Ofli, Weber,
  and Torralba}{Salvador et~al\mbox{.}}{2017}]%
        {salvador2017learning}
\bibfield{author}{\bibinfo{person}{Amaia Salvador}, \bibinfo{person}{Nicholas
  Hynes}, \bibinfo{person}{Yusuf Aytar}, \bibinfo{person}{Javier Marin},
  \bibinfo{person}{Ferda Ofli}, \bibinfo{person}{Ingmar Weber}, {and}
  \bibinfo{person}{Antonio Torralba}.} \bibinfo{year}{2017}\natexlab{}.
\newblock \showarticletitle{Learning cross-modal embeddings for cooking recipes
  and food images}. In \bibinfo{booktitle}{\emph{Proceedings of the IEEE
  conference on computer vision and pattern recognition}}.
  \bibinfo{pages}{3020--3028}.
\newblock


\bibitem[\protect\citeauthoryear{Saseendran, Skubch, and Keuper}{Saseendran
  et~al\mbox{.}}{2021}]%
        {saseendran2021multi}
\bibfield{author}{\bibinfo{person}{Amrutha Saseendran},
  \bibinfo{person}{Kathrin Skubch}, {and} \bibinfo{person}{Margret Keuper}.}
  \bibinfo{year}{2021}\natexlab{}.
\newblock \showarticletitle{Multi-Class Multi-Instance Count Conditioned
  Adversarial Image Generation}. In \bibinfo{booktitle}{\emph{Proceedings of
  the IEEE/CVF International Conference on Computer Vision}}.
  \bibinfo{pages}{6762--6771}.
\newblock


\bibitem[\protect\citeauthoryear{Sharma, Suhubdy, Michalski, Kahou, and
  Bengio}{Sharma et~al\mbox{.}}{2018}]%
        {sharma2018chatpainter}
\bibfield{author}{\bibinfo{person}{Shikhar Sharma}, \bibinfo{person}{Dendi
  Suhubdy}, \bibinfo{person}{Vincent Michalski},
  \bibinfo{person}{Samira~Ebrahimi Kahou}, {and} \bibinfo{person}{Yoshua
  Bengio}.} \bibinfo{year}{2018}\natexlab{}.
\newblock \showarticletitle{Chatpainter: Improving text to image generation
  using dialogue}.
\newblock \bibinfo{journal}{\emph{arXiv preprint arXiv:1802.08216}}
  (\bibinfo{year}{2018}).
\newblock


\bibitem[\protect\citeauthoryear{Srivastava and Salakhutdinov}{Srivastava and
  Salakhutdinov}{2012}]%
        {srivastava2012learning}
\bibfield{author}{\bibinfo{person}{Nitish Srivastava} {and}
  \bibinfo{person}{Ruslan Salakhutdinov}.} \bibinfo{year}{2012}\natexlab{}.
\newblock \showarticletitle{Learning representations for multimodal data with
  deep belief nets}. In \bibinfo{booktitle}{\emph{International conference on
  machine learning workshop}}, Vol.~\bibinfo{volume}{79}.
  \bibinfo{pages}{978--1}.
\newblock


\bibitem[\protect\citeauthoryear{Sukhbaatar, Weston, Fergus,
  et~al\mbox{.}}{Sukhbaatar et~al\mbox{.}}{2015}]%
        {sukhbaatar2015end}
\bibfield{author}{\bibinfo{person}{Sainbayar Sukhbaatar},
  \bibinfo{person}{Jason Weston}, \bibinfo{person}{Rob Fergus},
  {et~al\mbox{.}}} \bibinfo{year}{2015}\natexlab{}.
\newblock \showarticletitle{End-to-end memory networks}.
\newblock \bibinfo{journal}{\emph{Advances in neural information processing
  systems}}  \bibinfo{volume}{28} (\bibinfo{year}{2015}).
\newblock


\bibitem[\protect\citeauthoryear{Sun, Li, Wang, Zhao, and Sun}{Sun
  et~al\mbox{.}}{2021}]%
        {sun2021multi}
\bibfield{author}{\bibinfo{person}{Jianxin Sun}, \bibinfo{person}{Qi Li},
  \bibinfo{person}{Weining Wang}, \bibinfo{person}{Jian Zhao}, {and}
  \bibinfo{person}{Zhenan Sun}.} \bibinfo{year}{2021}\natexlab{}.
\newblock \showarticletitle{Multi-caption Text-to-Face Synthesis: Dataset and
  Algorithm}. In \bibinfo{booktitle}{\emph{Proceedings of the 29th ACM
  International Conference on Multimedia}}. \bibinfo{pages}{2290--2298}.
\newblock


\bibitem[\protect\citeauthoryear{Szegedy, Vanhoucke, Ioffe, Shlens, and
  Wojna}{Szegedy et~al\mbox{.}}{2016}]%
        {szegedy2016rethinking}
\bibfield{author}{\bibinfo{person}{Christian Szegedy}, \bibinfo{person}{Vincent
  Vanhoucke}, \bibinfo{person}{Sergey Ioffe}, \bibinfo{person}{Jon Shlens},
  {and} \bibinfo{person}{Zbigniew Wojna}.} \bibinfo{year}{2016}\natexlab{}.
\newblock \showarticletitle{Rethinking the inception architecture for computer
  vision}. In \bibinfo{booktitle}{\emph{Proceedings of the IEEE conference on
  computer vision and pattern recognition}}. \bibinfo{pages}{2818--2826}.
\newblock


\bibitem[\protect\citeauthoryear{Tan, Liu, Li, Zhang, and Yin}{Tan
  et~al\mbox{.}}{2019}]%
        {tan2019semantics}
\bibfield{author}{\bibinfo{person}{Hongchen Tan}, \bibinfo{person}{Xiuping
  Liu}, \bibinfo{person}{Xin Li}, \bibinfo{person}{Yi Zhang}, {and}
  \bibinfo{person}{Baocai Yin}.} \bibinfo{year}{2019}\natexlab{}.
\newblock \showarticletitle{Semantics-enhanced adversarial nets for
  text-to-image synthesis}. In \bibinfo{booktitle}{\emph{Proceedings of the
  IEEE/CVF International Conference on Computer Vision}}.
  \bibinfo{pages}{10501--10510}.
\newblock


\bibitem[\protect\citeauthoryear{Tao, Tang, Wu, Sebe, Jing, Wu, and Bao}{Tao
  et~al\mbox{.}}{2020}]%
        {tao2020df}
\bibfield{author}{\bibinfo{person}{Ming Tao}, \bibinfo{person}{Hao Tang},
  \bibinfo{person}{Songsong Wu}, \bibinfo{person}{Nicu Sebe},
  \bibinfo{person}{Xiao-Yuan Jing}, \bibinfo{person}{Fei Wu}, {and}
  \bibinfo{person}{Bingkun Bao}.} \bibinfo{year}{2020}\natexlab{}.
\newblock \showarticletitle{Df-gan: Deep fusion generative adversarial networks
  for text-to-image synthesis}.
\newblock \bibinfo{journal}{\emph{arXiv preprint arXiv:2008.05865}}
  (\bibinfo{year}{2020}).
\newblock


\bibitem[\protect\citeauthoryear{Wang, Lin, Hoi, and Miao}{Wang
  et~al\mbox{.}}{2021}]%
        {wang2021cycle}
\bibfield{author}{\bibinfo{person}{Hao Wang}, \bibinfo{person}{Guosheng Lin},
  \bibinfo{person}{Steven~CH Hoi}, {and} \bibinfo{person}{Chunyan Miao}.}
  \bibinfo{year}{2021}\natexlab{}.
\newblock \showarticletitle{Cycle-consistent inverse gan for text-to-image
  synthesis}. In \bibinfo{booktitle}{\emph{Proceedings of the 29th ACM
  International Conference on Multimedia}}. \bibinfo{pages}{630--638}.
\newblock


\bibitem[\protect\citeauthoryear{Welinder, Branson, Mita, Wah, Schroff,
  Belongie, and Perona}{Welinder et~al\mbox{.}}{2010}]%
        {welinder2010caltech}
\bibfield{author}{\bibinfo{person}{Peter Welinder}, \bibinfo{person}{Steve
  Branson}, \bibinfo{person}{Takeshi Mita}, \bibinfo{person}{Catherine Wah},
  \bibinfo{person}{Florian Schroff}, \bibinfo{person}{Serge Belongie}, {and}
  \bibinfo{person}{Pietro Perona}.} \bibinfo{year}{2010}\natexlab{}.
\newblock \showarticletitle{Caltech-UCSD birds 200}.
\newblock  (\bibinfo{year}{2010}).
\newblock


\bibitem[\protect\citeauthoryear{Wu, Liang, Ji, Yang, Fang, Jiang, and Duan}{Wu
  et~al\mbox{.}}{2021a}]%
        {wu2021n}
\bibfield{author}{\bibinfo{person}{Chenfei Wu}, \bibinfo{person}{Jian Liang},
  \bibinfo{person}{Lei Ji}, \bibinfo{person}{Fan Yang},
  \bibinfo{person}{Yuejian Fang}, \bibinfo{person}{Daxin Jiang}, {and}
  \bibinfo{person}{Nan Duan}.} \bibinfo{year}{2021}\natexlab{a}.
\newblock \showarticletitle{N$\backslash$" uwa: Visual synthesis pre-training
  for neural visual world creation}.
\newblock \bibinfo{journal}{\emph{arXiv preprint arXiv:2111.12417}}
  (\bibinfo{year}{2021}).
\newblock


\bibitem[\protect\citeauthoryear{Wu, Zhang, Wu, Wang, Li, Sun, and Li}{Wu
  et~al\mbox{.}}{2021c}]%
        {wu2021f3a}
\bibfield{author}{\bibinfo{person}{Xintian Wu}, \bibinfo{person}{Qihang Zhang},
  \bibinfo{person}{Yiming Wu}, \bibinfo{person}{Huanyu Wang},
  \bibinfo{person}{Songyuan Li}, \bibinfo{person}{Lingyun Sun}, {and}
  \bibinfo{person}{Xi Li}.} \bibinfo{year}{2021}\natexlab{c}.
\newblock \showarticletitle{F$^3$A-GAN: Facial Flow for Face Animation With
  Generative Adversarial Networks}.
\newblock \bibinfo{journal}{\emph{IEEE Transactions on Image Processing}}
  \bibinfo{volume}{30} (\bibinfo{year}{2021}), \bibinfo{pages}{8658--8670}.
\newblock


\bibitem[\protect\citeauthoryear{Wu, Wu, Tian, and Li}{Wu
  et~al\mbox{.}}{2021b}]%
        {wu2021mgh}
\bibfield{author}{\bibinfo{person}{Yiming Wu}, \bibinfo{person}{Xintian Wu},
  \bibinfo{person}{Jian Tian}, {and} \bibinfo{person}{Xi Li}.}
  \bibinfo{year}{2021}\natexlab{b}.
\newblock \showarticletitle{MGH: Metadata Guided Hypergraph Modeling for
  Unsupervised Person Re-identification}. \bibinfo{publisher}{{ACM}}.
\newblock


\bibitem[\protect\citeauthoryear{Xia, Zhang, Yang, Xue, Zhou, and Yang}{Xia
  et~al\mbox{.}}{2021}]%
        {xia2021gan}
\bibfield{author}{\bibinfo{person}{Weihao Xia}, \bibinfo{person}{Yulun Zhang},
  \bibinfo{person}{Yujiu Yang}, \bibinfo{person}{Jing-Hao Xue},
  \bibinfo{person}{Bolei Zhou}, {and} \bibinfo{person}{Ming-Hsuan Yang}.}
  \bibinfo{year}{2021}\natexlab{}.
\newblock \showarticletitle{GAN inversion: A survey}.
\newblock \bibinfo{journal}{\emph{arXiv preprint arXiv:2101.05278}}
  (\bibinfo{year}{2021}).
\newblock


\bibitem[\protect\citeauthoryear{Xiao, Chen, Zhang, Ji, Shao, Ye, and
  Xiao}{Xiao et~al\mbox{.}}{2021}]%
        {xiao2021boundary}
\bibfield{author}{\bibinfo{person}{Shaoning Xiao}, \bibinfo{person}{Long Chen},
  \bibinfo{person}{Songyang Zhang}, \bibinfo{person}{Wei Ji},
  \bibinfo{person}{Jian Shao}, \bibinfo{person}{Lu Ye}, {and}
  \bibinfo{person}{Jun Xiao}.} \bibinfo{year}{2021}\natexlab{}.
\newblock \showarticletitle{Boundary proposal network for two-stage natural
  language video localization}. In \bibinfo{booktitle}{\emph{Proceedings of the
  AAAI Conference on Artificial Intelligence}}, Vol.~\bibinfo{volume}{35}.
  \bibinfo{pages}{2986--2994}.
\newblock


\bibitem[\protect\citeauthoryear{Xu, Wang, Chen, and Li}{Xu
  et~al\mbox{.}}{2015}]%
        {xu2015empirical}
\bibfield{author}{\bibinfo{person}{Bing Xu}, \bibinfo{person}{Naiyan Wang},
  \bibinfo{person}{Tianqi Chen}, {and} \bibinfo{person}{Mu Li}.}
  \bibinfo{year}{2015}\natexlab{}.
\newblock \showarticletitle{Empirical evaluation of rectified activations in
  convolutional network}.
\newblock \bibinfo{journal}{\emph{arXiv preprint arXiv:1505.00853}}
  (\bibinfo{year}{2015}).
\newblock


\bibitem[\protect\citeauthoryear{Xu, Zhang, Huang, Zhang, Gan, Huang, and
  He}{Xu et~al\mbox{.}}{2018}]%
        {xu2018attngan}
\bibfield{author}{\bibinfo{person}{Tao Xu}, \bibinfo{person}{Pengchuan Zhang},
  \bibinfo{person}{Qiuyuan Huang}, \bibinfo{person}{Han Zhang},
  \bibinfo{person}{Zhe Gan}, \bibinfo{person}{Xiaolei Huang}, {and}
  \bibinfo{person}{Xiaodong He}.} \bibinfo{year}{2018}\natexlab{}.
\newblock \showarticletitle{Attngan: Fine-grained text to image generation with
  attentional generative adversarial networks}. In
  \bibinfo{booktitle}{\emph{Proceedings of the IEEE conference on computer
  vision and pattern recognition}}. \bibinfo{pages}{1316--1324}.
\newblock


\bibitem[\protect\citeauthoryear{Yang, Feng, Ji, Wang, and Chua}{Yang
  et~al\mbox{.}}{2021}]%
        {yang2021deconfounded}
\bibfield{author}{\bibinfo{person}{Xun Yang}, \bibinfo{person}{Fuli Feng},
  \bibinfo{person}{Wei Ji}, \bibinfo{person}{Meng Wang}, {and}
  \bibinfo{person}{Tat-Seng Chua}.} \bibinfo{year}{2021}\natexlab{}.
\newblock \showarticletitle{Deconfounded video moment retrieval with causal
  intervention}. In \bibinfo{booktitle}{\emph{Proceedings of the 44th
  International ACM SIGIR Conference on Research and Development in Information
  Retrieval}}. \bibinfo{pages}{1--10}.
\newblock


\bibitem[\protect\citeauthoryear{Ye, Yang, Takac, Sunderraman, and Ji}{Ye
  et~al\mbox{.}}{2021}]%
        {ye2021improving}
\bibfield{author}{\bibinfo{person}{Hui Ye}, \bibinfo{person}{Xiulong Yang},
  \bibinfo{person}{Martin Takac}, \bibinfo{person}{Rajshekhar Sunderraman},
  {and} \bibinfo{person}{Shihao Ji}.} \bibinfo{year}{2021}\natexlab{}.
\newblock \showarticletitle{Improving text-to-image synthesis using contrastive
  learning}.
\newblock \bibinfo{journal}{\emph{arXiv preprint arXiv:2107.02423}}
  (\bibinfo{year}{2021}).
\newblock


\bibitem[\protect\citeauthoryear{Yin, Liu, Sheng, Yu, Wang, and Shao}{Yin
  et~al\mbox{.}}{2019}]%
        {yin2019semantics}
\bibfield{author}{\bibinfo{person}{Guojun Yin}, \bibinfo{person}{Bin Liu},
  \bibinfo{person}{Lu Sheng}, \bibinfo{person}{Nenghai Yu},
  \bibinfo{person}{Xiaogang Wang}, {and} \bibinfo{person}{Jing Shao}.}
  \bibinfo{year}{2019}\natexlab{}.
\newblock \showarticletitle{Semantics disentangling for text-to-image
  generation}. In \bibinfo{booktitle}{\emph{Proceedings of the IEEE/CVF
  conference on computer vision and pattern recognition}}.
  \bibinfo{pages}{2327--2336}.
\newblock


\bibitem[\protect\citeauthoryear{Zhang, Koh, Baldridge, Lee, and Yang}{Zhang
  et~al\mbox{.}}{2021}]%
        {zhang2021cross}
\bibfield{author}{\bibinfo{person}{Han Zhang}, \bibinfo{person}{Jing~Yu Koh},
  \bibinfo{person}{Jason Baldridge}, \bibinfo{person}{Honglak Lee}, {and}
  \bibinfo{person}{Yinfei Yang}.} \bibinfo{year}{2021}\natexlab{}.
\newblock \showarticletitle{Cross-modal contrastive learning for text-to-image
  generation}. In \bibinfo{booktitle}{\emph{Proceedings of the IEEE/CVF
  Conference on Computer Vision and Pattern Recognition}}.
  \bibinfo{pages}{833--842}.
\newblock


\bibitem[\protect\citeauthoryear{Zhang, Xu, Li, Zhang, Wang, Huang, and
  Metaxas}{Zhang et~al\mbox{.}}{2017}]%
        {zhang2017stackgan}
\bibfield{author}{\bibinfo{person}{Han Zhang}, \bibinfo{person}{Tao Xu},
  \bibinfo{person}{Hongsheng Li}, \bibinfo{person}{Shaoting Zhang},
  \bibinfo{person}{Xiaogang Wang}, \bibinfo{person}{Xiaolei Huang}, {and}
  \bibinfo{person}{Dimitris~N Metaxas}.} \bibinfo{year}{2017}\natexlab{}.
\newblock \showarticletitle{Stackgan: Text to photo-realistic image synthesis
  with stacked generative adversarial networks}. In
  \bibinfo{booktitle}{\emph{Proceedings of the IEEE international conference on
  computer vision}}. \bibinfo{pages}{5907--5915}.
\newblock


\bibitem[\protect\citeauthoryear{Zhang, Xu, Li, Zhang, Wang, Huang, and
  Metaxas}{Zhang et~al\mbox{.}}{2018b}]%
        {zhang2018stackgan++}
\bibfield{author}{\bibinfo{person}{Han Zhang}, \bibinfo{person}{Tao Xu},
  \bibinfo{person}{Hongsheng Li}, \bibinfo{person}{Shaoting Zhang},
  \bibinfo{person}{Xiaogang Wang}, \bibinfo{person}{Xiaolei Huang}, {and}
  \bibinfo{person}{Dimitris~N Metaxas}.} \bibinfo{year}{2018}\natexlab{b}.
\newblock \showarticletitle{Stackgan++: Realistic image synthesis with stacked
  generative adversarial networks}.
\newblock \bibinfo{journal}{\emph{IEEE transactions on pattern analysis and
  machine intelligence}} \bibinfo{volume}{41}, \bibinfo{number}{8}
  (\bibinfo{year}{2018}), \bibinfo{pages}{1947--1962}.
\newblock


\bibitem[\protect\citeauthoryear{Zhang, Xie, and Yang}{Zhang
  et~al\mbox{.}}{2018a}]%
        {zhang2018photographic}
\bibfield{author}{\bibinfo{person}{Zizhao Zhang}, \bibinfo{person}{Yuanpu Xie},
  {and} \bibinfo{person}{Lin Yang}.} \bibinfo{year}{2018}\natexlab{a}.
\newblock \showarticletitle{Photographic text-to-image synthesis with a
  hierarchically-nested adversarial network}. In
  \bibinfo{booktitle}{\emph{Proceedings of the IEEE conference on computer
  vision and pattern recognition}}. \bibinfo{pages}{6199--6208}.
\newblock


\bibitem[\protect\citeauthoryear{Zhao, Wang, Fu, Wu, and Li}{Zhao
  et~al\mbox{.}}{2021}]%
        {zhao2021memory}
\bibfield{author}{\bibinfo{person}{Hanbin Zhao}, \bibinfo{person}{Hui Wang},
  \bibinfo{person}{Yongjian Fu}, \bibinfo{person}{Fei Wu}, {and}
  \bibinfo{person}{Xi Li}.} \bibinfo{year}{2021}\natexlab{}.
\newblock \showarticletitle{Memory efficient class-incremental learning for
  image classification}.
\newblock \bibinfo{journal}{\emph{IEEE Transactions on Neural Networks and
  Learning Systems}} (\bibinfo{year}{2021}).
\newblock


\bibitem[\protect\citeauthoryear{Zhu and Ngo}{Zhu and Ngo}{2020}]%
        {zhu2020cookgan}
\bibfield{author}{\bibinfo{person}{Bin Zhu} {and} \bibinfo{person}{Chong-Wah
  Ngo}.} \bibinfo{year}{2020}\natexlab{}.
\newblock \showarticletitle{CookGAN: Causality based text-to-image synthesis}.
  In \bibinfo{booktitle}{\emph{Proceedings of the IEEE/CVF Conference on
  Computer Vision and Pattern Recognition}}. \bibinfo{pages}{5519--5527}.
\newblock


\bibitem[\protect\citeauthoryear{Zhu, Pan, Chen, and Yang}{Zhu
  et~al\mbox{.}}{2019}]%
        {zhu2019dm}
\bibfield{author}{\bibinfo{person}{Minfeng Zhu}, \bibinfo{person}{Pingbo Pan},
  \bibinfo{person}{Wei Chen}, {and} \bibinfo{person}{Yi Yang}.}
  \bibinfo{year}{2019}\natexlab{}.
\newblock \showarticletitle{Dm-gan: Dynamic memory generative adversarial
  networks for text-to-image synthesis}. In
  \bibinfo{booktitle}{\emph{Proceedings of the IEEE/CVF Conference on Computer
  Vision and Pattern Recognition}}. \bibinfo{pages}{5802--5810}.
\newblock


\end{thebibliography}

\end{document}